\documentclass[sigconf]{acmart}
\AtBeginDocument{%
  }

\usepackage{ragged2e} 
\usepackage{booktabs,makecell, multirow, tabularx}
\usepackage{graphicx}
\usepackage{subfigure}

\newcommand{\name}{GraMI}

\begin{document}

\title{Variational Graph {Autoencoder} for Heterogeneous Information Networks with Missing and Inaccurate Attributes}

\author{Yige Zhao}
\email{ygzhao@stu.ecnu.edu.cn}
\affiliation{%
  \institution{East China Normal University}
  \city{Shanghai}
  \country{China}
}
\author{Jianxiang Yu}
\email{jianxiangyu@stu.ecnu.edu.cn}
\affiliation{%
  \institution{East China Normal University}
  \city{Shanghai}
  \country{China}
}
\author{Yao Cheng}
\email{yaocheng_623@stu.ecnu.edu.cn}
\affiliation{%
  \institution{East China Normal University}
  \city{Shanghai}
  \country{China}
}
\author{Chengcheng Yu}
\email{ccyu@sspu.edu.cn}
\affiliation{%
  \institution{Shanghai Polytechnic University}
  \city{Shanghai}
  \country{China}
}
\author{Yiding Liu}
\email{liuyiding.tanh@gmail.com}
\affiliation{%
  \institution{Baidu Inc.}
  \city{Beijing}
  \country{China}
}
\author{Xiang Li}
\email{xiangli@dase.ecnu.edu.cn}
\affiliation{%
  \institution{East China Normal University}
  \city{Shanghai}
  \country{China}
}
\author{Shuaiqiang Wang}
\email{shqiang.wang@gmail.com}
\affiliation{%
  \institution{Baidu Inc.}
  \city{Beijing}
  \country{China}
}
\renewcommand{\shortauthors}{Yige Zhao et al.}
\begin{abstract}
Heterogeneous Information Networks (HINs), which consist of various types of nodes and edges, have recently witnessed excellent performance in graph mining. However, most existing heterogeneous graph neural networks (HGNNs) fail to simultaneously handle 
the problems of missing attributes, inaccurate attributes and scarce node labels,
which limits their expressiveness. 
In this paper, we propose a generative self-supervised model \name\ to address these issues simultaneously. 
Specifically, 
\name\ first initializes all the nodes in the graph with a low-dimensional representation matrix.
After that,
based on the variational graph autoencoder framework,
\name\ learns both node-level and attribute-level embeddings in the encoder, 
which can provide fine-grained semantic information to construct node attributes. 
In the decoder,
\name\ reconstructs both links and attributes. 
Instead of directly reconstructing raw features for attributed nodes,
\name\ generates the initial low-dimensional representation matrix for all the nodes,
based on which raw features of attributed nodes are further reconstructed.
In this way,
\name\ can not only complete informative features for non-attributed nodes, 
but rectify inaccurate ones for attributed nodes.
Finally,
we conduct extensive experiments
to show the superiority of \name\ in tackling HINs with missing and inaccurate attributes.
Our code and data can be found here: \url{https://anonymous.4open.science/r/GraMI-82C9}.

\end{abstract}




\maketitle
\section{Introduction}
Heterogeneous Information Networks (HINs) are a type of information networks that incorporate various types of nodes and edges. 
In real-world scenarios,
HINs can effectively model data complexity, 
which provide rich semantics and a comprehensive view of data. 
Recently, Heterogeneous Graph Neural Networks (HGNNs) have received great attention and widely used in many related fields, such as social networks~\cite{hamilton2017inductive,wang2016structural}, recommender systems~\cite{berg2017graph,zhang2019star}, and knowledge graphs~\cite{sun2019rotate}. 
To perform an in-depth analysis on HINs, 
many HGNN models~\cite{zhang2019heterogeneous,wang2019heterogeneous,fu2020magnn} have been proposed to learn nodes' representations and perform well on downstream tasks like node classifcation~\cite{dong2020heterogeneous,yun2019graph} and link prediction~\cite{zhang2019heterogeneous,fu2020magnn}.

\textbf{Dilemmas.} 
At present, although heterogeneous graphs have received wide attention~\cite{li2021leveraging,sun2011pathsim1,hu2020heterogeneous,yang2023simple,mao2023hinormer}, there are two major challenges that are easily overlooked in most methods:

First, node attributes are generally incomplete in raw datasets. 
Collecting the attributes of all nodes is difficult due to the high cost and privacy concerns~\cite{zhu2023autoac}. 
Take the benchmark ACM dataset~\cite{jin2021heterogeneous} as an example: the heterogeneous graph modeled by ACM consists of three types of nodes: \emph{Paper}, \emph{Author} and~\emph{Subject}. The attributes of a paper node are derived from the keywords in its title, 
while the other two types of nodes lack attributes. 
Recent research has shown that the features of authors and subjects play a crucial role in learning the embeddings of heterogeneous graphs~\cite{fu2020magnn}. 
Hence, the completion of missing attributes is a matter of concern.

Second, inaccurate node attributes can lead to the spread of misinformation, 
which adversely affects the model performance.
In datasets such as ACM, attributes of \emph{Paper} nodes are typically extracted from bag-of-words representation of their keywords. However,
there might exist some noise. For example, some words that do not help express the topic may be included, or certain words might be mislabeled. 
According to the message passing mechanism of GNNs~\cite{kipf2016semi}, the representation of a node is obtained by aggregating information from its neighbors. 
If raw node attributes are inaccurate, 
the noise will propagate to the node's neighbors
and degrade the model's performance.
Therefore, 
it is important to alleviate the effect of inaccurate attributes in the graph.

Recently,
self-supervised learning (SSL), 
which attempts to extract information from the data itself, becomes a promising solution when no or few labels are provided~\cite{tian2022heterogeneous}. 
In particular,
generative
SSL~\cite{tian2022heterogeneous,liu2022graph,wu2021self} that aims to reconstruct the input graph
has been less studied in HINs.
Meanwhile,
existing models~\cite{he2022analyzing,tian2022heterogeneous,jin2021heterogeneous,xu2022graph,zhu2021graph} can only address either problem mentioned above.
Due to the prevalence of \emph{attribute incompleteness}, \emph{attribute inaccuracy} and 
\emph{label scarcity} in HINs,
there arises a question:
\emph{Can we develop an unsupervised generative model to jointly tackle the problem of missing and inaccurate attributes in HINs?}

To address the problem,
in this paper, we propose a variational \textbf{Gra}ph autoencoder for heterogeneous information networks with \textbf{M}issing and \textbf{I}naccurate attributes, namely, \name.  
As a generative model,
\name\ is unsupervised and does not rely on node labels in model training.
To deal with the problem of missing and inaccurate attributes,
\name\ first maps all the nodes in HINs, 
including both attributed and non-attributed ones, 
into the same low-dimensional space and generates a node representation matrix,
where
each row in the matrix corresponds to a node.
The low-dimensional representations can not only retain useful information and reduce noise in raw features of attributed nodes,
but also construct initial features for non-attributed ones.
After that,
\name\ learns both node and attribute embeddings by encoders and reconstructs both links and attributes by decoders. 
In particular, 
we learn embeddings of attributes in the low-dimensional space but not raw features.
On the one hand, by collaboratively generating node-level and attribute-level embeddings, fine-grained semantic information can be obtained for generating attributes. On the other hand, unlike most existing methods~\cite{li2023seegera,meng2019co} that directly reconstruct raw high-dimensional node features, \name\ instead reconstructs the low-dimensional node representation matrix. This approach not only alleviates the adverse effect of noise contained in raw node features, 
but also enhances the feature information for non-attribute nodes. 
Further, 
for attributed nodes,
we generate their raw features to leverage the information of accurate attributes.
In this way,
we can not only construct informative features for non-attributed nodes, 
but rectify inaccurate ones for attributed nodes.
Finally,
our main contributions are summarized as follows:
\begin{itemize}
    \item We propose
    a self-supervised heterogeneous graph auto-encoder \name. 
    To our knowledge,
    \name\ is the first self-supervised model that tackles the problems of attribute incompleteness and attribute inaccuracy in HINs.
    \item We present a novel feature reconstruction method for both attributed and non-attributed nodes,
    which can generate informative attributes for non-attributed nodes and rectify inaccurate attributes for attributed nodes.
    \item We conduct extensive experiments to evaluate the performance of \name. 
    The results show that \name\ surpasses other competitors in various downstream tasks.
\end{itemize}

\section{Related Work}
\subsection{Heterogeneous Graph Neural Networks}
HGNNs have recently attracted wide attention and 
existing models can be divided into two categories based on whether they employ meta-paths ~\cite{shi2016survey,zhang2019heterogeneous}. 
Some models ~\cite{wang2019heterogeneous,hu2020heterogeneous,fu2020magnn,sun2011pathsim} use meta-paths to capture high-order semantics in the graph. 
For example, 
HAN~\cite{wang2019heterogeneous} introduces a hierarchical attention mechanism, including node-level attention and semantic-level attention, to learn node embeddings. 
Both
MAGNN~\cite{fu2020magnn} 
and ConCH~\cite{li2021leveraging}
further consider intermediate nodes in instances of meta-paths to obtain more semantic information. 
There also exist some models ~\cite{schlichtkrull2018modeling,busbridge2019relational,yun2019graph} 
that do not require meta-paths. 
A representative model is 
HGT~\cite{hu2020heterogeneous},
a heterogeneous graph transformer, 
distinguishes the types of a node's neighbors and 
aggregates information from the neighbors according to node types. 
Further,
GTN~\cite{yun2019graph} learns a soft selection of edge types and {composite relations} for generating useful multi-hop connections. 

\subsection{Graph Learning with Missing/Inaccurate Attributes}
Due to privacy protection, 
missing attributes are ubiquitous in graph-structured data. 
Some recent methods have been proposed to solve the issue in homogeneous graphs, including 
{GRAPE}~\cite{you2020handling},
GCNMF~\cite{taguchi2021graph} and Feature Propagation~\cite{rossi2022unreasonable}. 
There are also methods~\cite{jin2021heterogeneous,he2022analyzing}
that are designed for HINs. For example,
HGNN-AC~\cite{jin2021heterogeneous} proposes the first framework to complete attributes for heterogeneous graphs,
which is based on the attention mechanism 
and only works in semi-supervised settings. 
HGCA~\cite{he2022analyzing} is an unsupervised learning method that attempts to tackle the missing attribute problem by contrastive learning.
Further,
graph-structured data in the real world is often  corrupted with noise, which leads to inaccurate attributes. 
Many methods~\cite{xu2022graph,zhu2021graph,wang2019graphdefense,zugner2020adversarial} have been proposed to enhance the model's robustness against noise by introducing graph augmentation, adversarial learning, and some other techniques on homogeneous graph. 
Due to HINs' complex structure, the impact of noise might be even more significant.

\subsection{Self-supervised Learning on Graphs}
Self-supervised learning on graphs can generally be divided into contrastive learning and generative approaches~\cite{hou2022graphmae,tian2022heterogeneous}. 
Specifically,
contrastive learning 
uses pre-designed data augmentation to obtain correlated views of a given graph and adopts a contrastive loss to learn representations that maximize the agreement between positive views while minimizing the disagreement between negative ones~\cite{he2022analyzing,velickovic2019deep,sun2019infograph,you2020graph}. 
Recently, HeCo~\cite{wang2021self}  
uses network schema and meta-paths as two views in HINs to align both local and global information of nodes. 
However, the success of contrastive learning heavily relies on informative data augmentation, 
which has been shown to be variant across datasets ~\cite{hou2022graphmae,tian2022heterogeneous}.

Generative learning aims to use the input graph for self-supervision and to recover the input data~\cite{li2023seegera}. 
In prior research, 
existing methods include those reconstructing only links~\cite{kipf2016variational,pan2018adversarially}, 
only features~\cite{hou2022graphmae}, 
or a combination of both links and features~\cite{salehi2019graph,meng2019co,li2023seegera,tian2022heterogeneous}. 
A recent model SeeGera~\cite{li2023seegera} proposes a hierarchical variational graph auto-encoder and achieves superior results on many downstream tasks. 
However, SeeGera is specially designed for homogeneous graphs and cannot be directly applied in HINs.
While some methods~\cite{tian2022heterogeneous,wang2021hgate} are presented for HINs, they are based on meta-paths. 
For example,
HGMAE~\cite{tian2022heterogeneous} adopts the masking mechanism 
and is proposed as a heterogeneous graph masked auto-encoder that generates both virtual links guided by meta-paths and features.
However, 
it ignores the problem of missing attributes in HINs.

\section{Preliminary}
\label{sec:pre}

\textbf{[Attributed Heterogeneous Information Networks (AHINs)]}.
An attributed heterogeneous information network (AHIN) is defined as a graph $\mathcal G=(\mathcal V,\mathcal E,\mathcal A),$ where $\mathcal V$ is the set of nodes, $\mathcal E$ is the set of edges and $\mathcal{A}$ is the set of node attributes.
Let $\mathcal{T} = \{T_1, \cdots, T_{\vert \mathcal{T} \vert}\}$ and $\mathcal{R} = \{r_1, \cdots, r_{\vert \mathcal{R} \vert}\}$ denote the node type set and edge type set, respectively.
Each node $v\in \mathcal V$ is associated with a node type by a mapping function $\varphi:\mathcal V\rightarrow \mathcal T$,
and each edge $e\in \mathcal E$ has an edge type with a mapping function $\phi:\mathcal E\rightarrow \mathcal R$.
When $|\mathcal T| = 1$ and $|\mathcal R| = 1$,
$\mathcal{G}$ reduces to a homogeneous graph.\\
\textbf{[AHINs with missing attributes]}.
Nodes in AHINs are usually associated with attributes.
Given a node type $T_i$,
we denote its corresponding attribute set as $\mathcal{A}_i \subset \mathcal{A}$.
In the real world,
it is prevalent that some node types in AHINs are given specific attributes while others' attributes are missing.
In this paper,
we divide $\mathcal{T}$ into two subsets: $\mathcal{T} = \mathcal{T}^{+} \cup \mathcal{T}^{-}$,
where $\mathcal{T}^{+}$ represents attributed node types
and $\mathcal{T}^{-}$ indicates non-attributed ones.\\
\textbf{[Variational Lower bound]}.
Given an HIN with the adjacency matrix $\mathrm A$ and the attribute matrix $\mathrm X$ as observations,
our goal is to learn both node embeddings $\mathrm Z^\mathcal V$
and attribute embeddings $\mathrm Z^\mathcal A$.
To approximate the true posterior distribution $p(\mathrm Z^\mathcal V,\mathrm Z^\mathcal A|\mathrm A,\mathrm X)$, 
following~\cite{li2023seegera}, 
we adopt semi-implicit variational inference that can capture a wide range of distributions more than Gaussian~\cite{yin2018semi}
and define a hierarchical variational distribution $h_{\phi}(\mathrm Z^\mathcal V,\mathrm Z^\mathcal A)$ with parameter $\phi$ to minimize $\textrm{KL}(h_{\phi}(\mathrm Z^\mathcal V,\mathrm Z^\mathcal A)||p(\mathrm Z^\mathcal V,\mathrm Z^\mathcal A \vert \mathrm A,\mathrm X))$,
which is equivalent to maximizing the ELBO~\cite{bishop2013variational}:
\begin{equation}
\nonumber
    \textrm{ELBO}=\mathrm E_{\mathrm Z^\mathcal V,\mathrm Z^\mathcal A\sim h_{\phi}(\mathrm Z^\mathcal V,\mathrm Z^\mathcal A)}[\log \frac{p(\mathrm Z^\mathcal V,\mathrm Z^\mathcal A,\mathrm A,\mathrm X)}{h_{\phi}(\mathrm Z^\mathcal V,\mathrm Z^\mathcal A)}]=\mathcal{L}.
    \label{elbo}
\end{equation}
From~\cite{li2023seegera}, 
we assume the independence between 
$\mathrm Z^\mathcal V$ and $\mathrm Z^\mathcal A$ for simplicity,
and derive a lower bound $\underline{\mathcal{L}}$ for the ELBO:
\begin{align}
    \underline{\mathcal{L}}
      = \ &\mathrm E_{\mathrm{Z}^\mathcal V\sim{h_{\phi_1}(\mathrm{Z}^\mathcal V)}}\log p(\mathrm A|\mathrm{Z}^\mathcal V)-\textrm{KL}(h_{\phi_1}(\mathrm{Z}^\mathcal V)||p(\mathrm{Z}^\mathcal V))\notag \\& 
      +\mathrm E_{\mathrm{Z}^\mathcal V\sim{h_{\phi_1}(\mathrm{Z}^\mathcal V),\mathrm{Z}^\mathcal A\sim h_{\phi_2}(\mathrm{Z}^\mathcal A)}}\log p(\mathrm X|\mathrm{Z}^\mathcal V,\mathrm{Z}^\mathcal A)\notag \\&-\textrm{KL}(h_{\phi_2}(\mathrm{Z}^\mathcal A)||p(\mathrm{Z}^\mathcal A))
      \label{ourselbo}
\end{align}
where  
$h_{\phi_1}(\mathrm Z^\mathcal V)$ and $h_{\phi_2}(\mathrm Z^\mathcal A)$ are variational distributions generated from the node encoder and the attribute encoder, respectively.
Further,
$ p(\mathrm A|\mathrm Z^\mathcal V)$ and $p(\mathrm X|\mathrm Z^\mathcal V,\mathrm Z^\mathcal A)$ are used to reconstruct 
both $\mathrm A$ and $\mathrm X$.
Note that 
deriving the lower bound under the independence or the correlation assumption is not the focus of this paper. 
Our proposed model can be easily adapted to the correlated case in~\cite{li2023seegera}.

\section{Algorithm}



\begin{figure*}[htbp]
    \centering
    \includegraphics[width=0.75\linewidth]{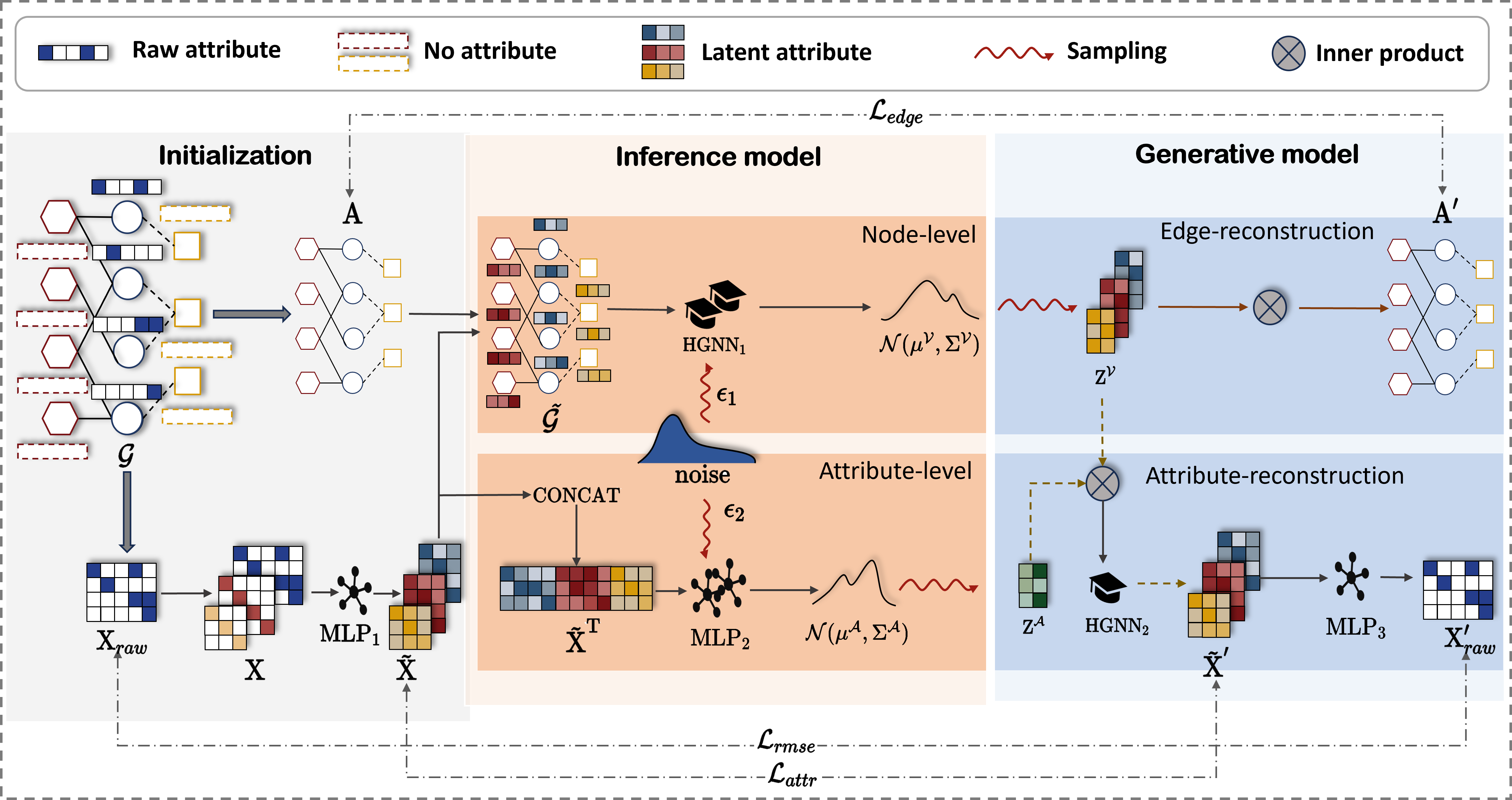}
    \caption{ The overall framework of \name\ }
    \label{fig:Tetris}
\end{figure*}

\subsection{Initialization}
HINs contain various types of nodes,
 where some of them could have no attributes. 
 For attributed nodes in type ${T}_i$, 
 we retain the raw feature matrices $\mathrm X_i \in \mathbb R^{n_i\times d_i}$; 
 for non-attributed nodes in type ${T}_j$,
 we use 
 the one-hot encoded matrix
 $\mathrm I_j\in \mathbb R^{n_j\times n_j}$ 
 to initialize the feature matrix $\mathrm X_j$, where $n_i$ and $n_j$ are the number of nodes in types ${T}_i$ and ${T}_j$, respectively. 
 Note that various types of nodes could have attributes in different dimensions and semantic spaces.
So we apply a type-specific linear transformation for each type of nodes to map their
feature vectors into the same latent space with dimensionality $\tilde{d}$. 
Specifically,
for a node $v$ of type $T$, 
we have:
\begin{equation}
   \tilde x_v = \texttt{tanh}(\textbf W_T\cdot x_v+\textbf b_T), 
\end{equation}
where $\textbf W_T\in \mathbb R^{\tilde{d}\times d}$, $\textbf{b}_T\in \mathbb R^{\tilde{d}}$ are learnable parameter matrices for node type $T$, $x_v\in \mathbb R^{d}$ is the raw feature vector and $\tilde x_v\in \mathbb R^{\tilde{d}}$ is the hidden representation vector of node $v$ with $\tilde{d}\ll d$.

\subsection{Inference model}
\label{sec:inference}

We first map both nodes and attributes into low-dimensional embeddings with an inference model,
which consists of a node-level encoder and an attribute-level encoder.

\textbf{Node-level encoder.} 
To generate embeddings $\mathrm Z^{\mathcal V_{i}}$ for nodes in type $T_i$,
we first assume 
$q_{1}(\mathrm Z^{\mathcal V_{i}}|\mathrm A,\mathrm {\tilde X})=\prod_{j=1}^{n_i}q_{1}(z_j^{\mathcal V_{i}}|\mathrm A,\mathrm {\tilde X})$ with $q_{1}  (z_j^{\mathcal V_{i}}|\mathrm A,\mathrm {\tilde X}) = 
\mathcal N[z_j^{\mathcal V_{i}}|\mu_j^{\mathcal V_i}, \Sigma_j^{\mathcal{V}_i}]$,
where $\mathcal{N}$ denotes multivariant Gaussian distribution with mean $\mu_j^{\mathcal V_i}$ and 
diagonal co-variance matrix $\Sigma_j^{\mathcal{V}_i}$,
and $\mathrm {\tilde X} = \texttt{CONCAT}(\mathrm {\tilde X}_1 \Vert \cdots \Vert \mathrm {\tilde X}_{\vert \mathcal{T} \vert})$
denotes the hidden representation matrix for all the nodes.
To model 
$\mu_j^{\mathcal V_i}$ and 
$\Sigma_j^{\mathcal{V}_i}$ as random variables,
according to the semi-implicit variational inference~\cite{hasanzadeh2019semi},
we inject random noise $\epsilon_1$
into $\mathrm {\tilde X}$,
and derive:
\begin{equation}
    [\mu_j^{\mathcal V_i}, \Sigma_j^{\mathcal{V}_i}] = \texttt{HGNN}(\mathrm A,\mathrm {\texttt{CONCAT}(\mathrm {\tilde X},\epsilon_1)}),\ \epsilon_1 \sim q_1(\epsilon),
\end{equation}
where 
$q_1(\epsilon)$ is a noise distribution and set to be standard Gaussian distribution in our experiments.
$\Sigma_j^{\mathcal{V}_i}$ is a diagonal matrix with the output vector of HGNN as its diagonal.
Note that $\texttt{HGNN}(\cdot)$
can theoretically be any heterogeneous graph neural network models.
However,
to broaden the model's applicability in more downstream tasks,
we aim to encode all the nodes in the graph and also avoid the limitation of pre-given meta-paths.
We thus adopt a simple HGNN model next.
Other advanced HGNN models could lead to better model performance, but not our focus.


For each node $u$ 
and its neighbor $v$ connected by relation $r$, we 
use the
$\texttt{softmax}$ function 
to calculate the attention weight $\alpha_{uv}^r$ by: 
\begin{equation}
    \alpha^r_{uv}=\texttt{softmax}(e^r_{uv})=\frac{\exp(e^r_{uv})}{\sum_{v'\in \mathcal N^r_u}\exp(e^r_{uv'})}.
\end{equation}
Here, 
$e^r_{uv}=a^r(\mathrm W^rx_u,\mathrm W^rx_v)$,
where 
$x_u$ and $x_v$ are 
feature vectors, $\mathrm W^r$
and $a^r$ are parameters to be learned,
and
$\mathcal N_u^r$ represents the first-order neighbors of node $u$ induced by $r$.
Then based on the learned weights,
we aggregate the information from neighbors $\mathcal N_u^r$ and generate
the embedding $h_u^r$ w.r.t. relation $r$.
To stabilize the learning process, 
we can further employ multi-head attention.
Finally, after obtaining all relation-specific representations $\{h^{r_1}_u...h^{r_m}_u\}$ for node $u$, 
we generate its final representation vector by:
\begin{equation}
    h_u = \texttt{MEAN}(\{h^r_u\}).
\end{equation}

\textbf{Attribute-level encoder.} 
To further extract knowledge from
node attributes,
inspired by~\cite{li2023seegera,meng2019co},
we also encode node attributes.
Different from existing methods that encode raw node attributes,
we propose to encode hidden node features in the low-dimensional space.
There are two reasons that account for this.
On the one hand, 
given numerous features (e.g., text tokens),
feature encoding could lead to expensive time and memory complexity.
On the other hand,
when raw node features are missing or inaccurate,
encoding these features could bring noise.
To generate attribute embeddings $\mathrm Z^{\mathcal A_i}$ for nodes of type $T_i$,
similar as in the node-level encoder,
we assume 
$q_{2}(\mathrm Z^{\mathcal A_i}|\mathrm {\tilde X^T}_i) = \prod_{l=1}^{\tilde{d}} q_{2}(z_l^{\mathcal A_i}|\mathrm {\tilde X^T}_i)$, 
$q_{2}(z_l^{\mathcal A_i}|\mathrm {\tilde X^T}_i) =\mathcal N[z_l^{\mathcal A_i}|\mu_l^{\mathcal{A}_i}, \Sigma_l^{\mathcal{A}_i}]$,
where $\mathcal{N}$ denotes 
multivariant Gaussian distribution with mean $\mu_l^{\mathcal {A}_i}$ and 
diagonal co-variance matrix $\Sigma_l^{\mathcal{A}_i}$.
To model $\mu_l^{\mathcal A_i}$ and 
$\Sigma_l^{\mathcal{A}_i}$ with semi-implicit VI,
we inject noise into the hidden node feature matrix $\mathrm {\tilde X}_i$ and have: 
\begin{equation}
    [\mu_l^{\mathcal A_i}, \Sigma_l^{\mathcal{A}_i}]=\texttt{MLP}(\texttt{CONCAT}(\mathrm {\tilde X^T}_i,\epsilon_2)),\ \epsilon_2 \sim q_2(\epsilon)
\end{equation}
where 
$q_2(\epsilon)$ is another noise distribution and also set to be standard Gaussian
distribution in our experiments.
We use $\mathrm {\tilde X^T}_i \in \mathbb R^{\tilde{d}\times n_i}$ here because
we consider the $l$-th column of 
feature matrix $\mathrm {\tilde X}_i$
as the feature vector of the $l$-th attribute.
\textbf{Note that 
we only learn embeddings for hidden node features, but not raw features}.

\subsection{Generative model}
The generative model is used to reconstruct both edges $\mathcal{E}$ and node attributes in a heterogeneous graph. 


\textbf{Edge reconstruction.} 
Since 
there exist multiple types of edges in an HIN,
we distinguish these edges based on two end nodes.
For each edge $\mathrm A_{uv}$,
we draw
$\mathrm A_{uv} \sim \texttt{Ber}(p_{uv})$,
where $\texttt{Ber}(\cdot)$ denotes Bernoulli distribution and 
$p_{uv}$
is the existence probability of the edge between nodes $x_u$
and $x_v$.
Specifically,
given nodes $x_u$ in type $T_i$ 
and $x_v$ in type $T_j$,
we have
$p_{uv} = p(\mathrm A_{uv}=1|z^{\mathcal V_i}_u,z^{\mathcal V_j}_v)
=\sigma((z^{\mathcal V_i}_u)^\mathrm T z^{\mathcal V_j}_v)$,
where $\sigma$ is the $\texttt{sigmoid}$ function.


\textbf{Attribute reconstruction.} 
Since nodes could have missing or inaccurate features,
directly reconstructing raw node attributes could bring noise.
To address the issue,
we propose to generate the hidden embedding matrix $\mathrm {\tilde X}$ instead,
which introduces three major benefits.
First,
$\mathrm {\tilde X}$ has smaller dimensionality than
the original feature matrix $\mathrm {X}$;
hence, 
reconstructing
$\mathrm {\tilde X}$ needs less computational cost.
Second,
$\mathrm {\tilde X}$ contains rich semantic information that can cover missing attributes.
Third,
when raw features are inaccurate,
$\mathrm {\tilde X}$ contains less noise than $\mathrm X$,
and reconstructing $\mathrm {\tilde X}$  can further remove noise due to the well-known denoising effects of auto-encoders.

Given a node type $T_i$,
let  
$\mathrm {\tilde X}_i^{'}$ be its corresponding \textbf{reconstructed} hidden representation matrix.
For any node $x_u$ of type $T_i$,
we first initialize its $j$-th embedding value ${\mathrm {\tilde X}_i^{'}}[u,j]$
as:
\begin{equation}
       \mathrm {\tilde X}_i^{'}[u,j] =\texttt{tanh}((z^\mathcal V_i)^\mathrm Tz^\mathcal A_j).
\end{equation}
After that, 
taking these initial matrices for all the node types as input,
we further feed them into a $\texttt{HGNN}$ model to 
generate $\mathrm {\tilde X}^{'} = \texttt{HGNN}(\mathrm{A}, \mathrm {\tilde X}^{'})$.
Note that for node types with missing attributes,
we only reconstruct the hidden representation matrices.
For nodes associated with features,
despite the noise,
they could also provide rich useful information. Therefore,
we further reconstruct the raw feature matrices.
Assume that nodes in type $T_i$ have raw features.
Then we generate the raw feature matrix by a MLP model:
\begin{equation}
    \mathrm X_i^{'}=\texttt{MLP}(\mathrm{\tilde X}_i^{'}).
\end{equation}
Finally, the 
overall framework of \name\ is given in Figure~\ref{fig:Tetris}.

\subsection{Optimization}
In section~\ref{sec:pre},
we have given 
the lower bound $\underline{\mathcal L}$ for ELBO in Eq.~\eqref{ourselbo}. 
We then generalize $\underline{\mathcal L}$
to HINs and get:
\begin{equation}
\nonumber
\small
    \begin{split}
    &\tilde{\underline{\mathcal{L}}}
    =\underbrace{-\sum_{r\in \mathcal{R}}\mathrm E_{{h_{\phi_1}(\mathrm{Z}^\mathcal V)}}\log p(\mathrm A^r|\mathrm{Z}^\mathcal V)+\text{KL}(h_{\phi_1}(\mathrm{Z}^\mathcal V)||p(\mathrm{Z}^\mathcal V))}_{\mathcal{L}_{edge}}\\
    &\underbrace{-\sum_{T_i \in \mathcal{T}}\mathrm E_{{h_{\phi_1}(\mathrm{Z}^\mathcal V),h_{\phi_2}(\mathrm{Z}^\mathcal A)}}\log p(\tilde{\mathrm X}_{i}|\mathrm{Z}^\mathcal V,\mathrm{Z}^\mathcal A)+\text{KL}(h_{\phi_2}(\mathrm{Z}^\mathcal A)||p(\mathrm{Z}^\mathcal A))}_{\mathcal{L}_{attr}},    
    \end{split}
\end{equation}
where $\mathrm A^r$ is the adjacency matrix of relation $r \in \mathcal R$ and $\tilde{\mathrm X}_{i}$ is the hidden representation matrix corresponding to node type $T_i$. 
Further,
following auto-encoder~\cite{kipf2016variational},
since raw node attributes contain informative knowledge,
we use
the Root Mean Squared Error (RMSE) loss to ensure the closeness between reconstructed feature matrices and raw ones.
Formally,
the objective is given as:
\begin{equation}
    \mathcal{L}_{rmse} =\sqrt{\frac{1}{\vert \mathcal{T}^{+}\vert}\sum_{T_i\in \mathcal{T}^{+}}\Vert\mathrm X^{'}_i- \mathrm X_i\Vert^2},
\end{equation}
which is used as a regularization term to facilitate model training.
Finally, we train our model with the objective:
\begin{equation}    \mathcal{{L}}_{all}=\mathcal{L}_{edge}+\lambda_1\mathcal{L}_{attr}+\lambda_2\mathcal{L}_{rmse}.
    \label{L1}
\end{equation}
Here,
we introduce two hyperparameters $\lambda_1$ and $\lambda_2$ to
control the importance of different terms.\\
\textbf{[Time Complexity analysis]}
The major time complexity in the encoder comes from HGNN and MLP for nodes and attributes, respectively.
We use the simple HGNN introduced in Sec.~\ref{sec:inference}.
Suppose for each type of nodes, 
they have an average number of $m$ related adjacency matrices $\{\mathrm A^r\}_{r=1}^m$.
Since adjacency matrix is generally sparse,
for each $\mathrm A^r$,
let 
$n_{\mathrm A^r}^{row}$, $n_{\mathrm A^r}^{col}$ and 
$d_{\mathrm A^r}$
be the average number of rows, columns and non-zero entries in each row, respectively.
Note that we use the hidden embedding matrix $\mathrm {\tilde X}$ as input whose dimensionality is $n \times \tilde{d}$.
For simplicity, 
we denote the embedding dimensionality as $k$ in hidden layers of both HGNN and MLP.
Further,
let $\check{d}$ and $\hat{d}$ be the dimensionalities of injected noise to HGNN and MLP, 
respectively.
Then,
the time complexities for HGCN and MLP are
$O (m(n_{\mathrm A^r}^{row}d_{\mathrm A^r}(\tilde{d}+\check{d}) + n_{\mathrm A^r}^{row}(\tilde{d}+\check{d})k))$
and
$O(\tilde{d}(n+\hat{d})k)$,
respectively.
Both time complexities are linear to the number of nodes $n$ in the HIN.
In the decoder,
to reconstruct attributes,
in addition to HGNN and MLP, 
we have an additional inner product operation with
time complexity of $O (n\tilde{d}k)$,
which ensures an overall linear time complexity w.r.t. $n$.
For link reconstruction,
the time complexity is $O(mn_{\mathrm A^r}^{row}n_{\mathrm A^r}^{col}k)$.
As suggested by~\cite{kipf2016variational},
we can down-sample the number of nonexistent edges in the graph to reduce the time complexity for recovering links.

\noindent
\textbf{[Space Complexity Analysis]}
For space complexity, we assume all HGNNs and MLPs are one-layer. We define the total number of nodes as $n$, the initial attribute dimension as $d$, the initial low-dimensional dimension as $\tilde{d}$, and the encoded dimension as $k$. For the inference model, the space complexity for HGNN and MLP are $O(\tilde{d}\times k)$ and $O(n\times k)$. For the generative model, the space complexity for HGNN and MLP are $O(\tilde{d}\times \tilde{d})$ and $O(\tilde{d}\times d)$. Therefore, the total time complexity of the model is $O(\tilde{d}\times k+n\times k+\tilde{d}\times \tilde{d}+\tilde{d}\times d)$, which is also linear to the number of nodes.

\subsection{Discussion}
We next summarize the main difference between \name\ and the SOTA generative model HGMAE for HINs.
Although both models adopt an encoder-decoder framework,
they differ in three main aspects.
First,
HGMAE cannot deal with the problem of missing attributes in HINs.
When node attributes are missing,
HGMAE estimates and fills in the attribute values in the pre-processing step.
When the estimated values are inaccurate, the model performance could degenerate.
However,
\name\ takes node attributes as learnable parameters
and generates low-dimensional attributes with the decoder.
The learning process ensures the high quality of reconstructed node attributes.
Second,
when node attributes are inaccurate,
the masking mechanism adopted by HGMAE can enhance the model's robustness against noise to some degree.
However,
our model \name\ 
essentially solves the problem
by rectifying incorrect features and reconstructing more accurate ones.
This further boosts the generalizability of our model.
Third,
HGMAE focuses on learning embeddings of nodes in target types only. 
This narrows the model's application on downstream tasks centering around nodes in target types.
In comparison,
\name\ collectively learns embeddings for all the nodes in the graph,
which broadens the applicability of our model.


\section{EXPERIMENTS}

\subsection{Experimental Settings}
\subsubsection{Datasets and Baselines.} 
We conduct experiments on four real-world HIN datasets:
ACM~\cite{wang2019heterogeneous}, DBLP~\cite{sun2012mining}, YELP~\cite{lu2019relation} and AMiner~\cite{hu2019adversarial}. 
In these datasets,
only \emph{paper} nodes in ACM and DBLP, and \emph{business} nodes in YELP have raw attributes. 
For AMiner, 
there are not attributes for all types of nodes. 

In the classification task,
we compare \name\ with 10 other semi-supervised/unsupervised baselines, 
including methods for homogeneous graphs: GAT~\cite{velivckovic2017graph},  
DGI~\cite{velivckovic2018deep}, SeeGera~\cite{li2023seegera};
methods for HINs: HAN~\cite{wang2019heterogeneous}, MAGNN~\cite{fu2020magnn}, MAGNN-AC~\cite{jin2021heterogeneous}, Mp2vec~\cite{dong2017metapath2vec}, DMGI~\cite{park2020deep}, HGCA~\cite{he2022analyzing} and HGMAE~\cite{tian2022heterogeneous}. 
In particular,
MAGNN-AC and HGCA are SOTA methods for attribute completion in HINs;
SeeGera and HGMAE are SOTA generative SSL models on graphs in node classification.
In the link prediction task,
we compare \name\  with four methods for homogeneous graphs, i.e., VGAE~\cite{kipf2016variational}, SIG-VAE~\cite{hasanzadeh2019semi}, CAN~\cite{meng2019co}, SeeGera~\cite{li2023seegera}, and three methods for HINs that can encode all types of nodes: RGCN~\cite{schlichtkrull2018modeling}, HGT~\cite{hu2020heterogeneous},
MHGCN~\cite{yu2022multiplex}.

\begin{table*}[t]
\centering
\caption{Node classification results (\%). 
We highlight the best score on each dataset in bold and underline the runner-up.
We also perform statistically significant test, where ** and * denote p-value $<0.01$ and $<0.05$, respectively.}
\label{table3}
\resizebox{0.82\linewidth}{!}{
\begin{tabular}{c||c|c|ccc|ccccccc}
\hline
\multirow{2}{*}{Dataset} & \multirow{2}{*}{Metric} & \multirow{2}{*}{Training} & \multicolumn{3}{c|}{Semi-supervised} & \multicolumn{7}{c}{Unsupervised}\\
\cline{4-5}\cline{6-13}
\multirow{2}{*}{} & \multirow{2}{*}{} & \multirow{2}{*}{} &GAT& HAN & MAGNN  &Mp2vec&DGI& DMGI & SeeGera & HGMAE& HGCA & \name\ \\
\hline
\multirow{8}{*}{ACM} & \multirow{5}{*}{Macro-F1} & 10\% &89.51& 90.32 & 88.82 &69.47&89.55& 91.61& 89.34 &85.21& \textbf{92.10} & \underline{91.62} \\
\multirow{8}{*}{} &  & 20\% &89.77& 90.71 & 89.41 &70.11&90.06& \underline{92.22} & 89.85 &85.75& \textbf{92.54} & 92.17 \\
\multirow{8}{*}{} & \multirow{4}{*}{} & 40\% &89.92& 91.33 & 89.83 &70.43&90.19& 92.51 & 90.21 &86.52&\underline{93.00} & \textbf{93.01} \\
\multirow{8}{*}{} & \multirow{4}{*}{} & 60\%  &90.07&91.73 & 90.18 &70.73&90.34& 92.79 & 90.06 &87.12& \underline{93.28} & \textbf{93.36} \\
\multirow{8}{*}{} & \multirow{4}{*}{} & 80\% &89.76& 91.91 & 90.11 &71.13&90.20& 92.57 & 90.29 &88.68& \underline{93.21} & \textbf{93.49}*\\
\cline{2-13}
\multirow{8}{*}{} & \multirow{5}{*}{Micro-F1} & 10\% &89.42&90.05 & 88.81&73.81&89.54& 91.49 & 89.45&85.94 & \textbf{92.01} & \underline{91.51} \\
\multirow{8}{*}{} & \multirow{4}{*}{} & 20\% &89.67&90.59 & 89.36 &74.44&89.92& 92.07 & 89.85&86.14 & \textbf{92.45} &\underline{92.08} \\
\multirow{8}{*}{} & \multirow{4}{*}{} & 40\% &89.83&91.22	& 89.81	&74.80&90.04& 92.37 & 90.21 &86.74& \underline{92.92} & \textbf{92.98} \\
\multirow{8}{*}{} & \multirow{4}{*}{} & 60\% &89.98& 91.60 & 90.11 &75.22&90.17& 92.61 & 90.03 &87.25& \underline{93.18} & \textbf{93.23} \\
\multirow{8}{*}{} & \multirow{4}{*}{} & 80\% &89.70& 91.76 & 90.06 &75.57&90.00& 92.38 & 90.34 &88.98&  \underline{93.03} &\textbf{ 93.38}** \\
\hline
\multirow{8}{*}{DBLP} & \multirow{5}{*}{Macro-F1} & 10\% &81.90& 92.33 & \underline{92.52} &74.82&68.92& 91.88 & 83.92 &88.54& 90.79 & \textbf{93.64}** \\
\multirow{8}{*}{} & \multirow{4}{*}{} & 20\% &82.20& 92.63 & \underline{92.70} &76.66&77.11& 92.24 & 84.01 &88.71& 92.28 & \textbf{94.02}**\\
\multirow{8}{*}{} & \multirow{4}{*}{} & 40\% &82.17& 92.87 & 92.69 &82.14&81.09& 92.50 & 85.12&89.33 & \underline{93.02} & \textbf{94.17}**\\
\multirow{8}{*}{} & \multirow{4}{*}{} & 60\% &82.12& 93.05 & 92.75 &84.25&82.17& 92.60 & 85.78&89.83 & \underline{93.25} & \textbf{94.48}**\\
\multirow{8}{*}{} & \multirow{4}{*}{} & 80\% &82.02& 93.16 & 93.01 &84.20&82.68& 92.88 & 86.00 &91.40& \underline{93.82} & \textbf{94.68}**\\
\cline{2-13}
\multirow{8}{*}{} & \multirow{5}{*}{Micro-F1} & 10\% &83.23& 92.97 & \underline{93.08} & 75.86 & 76.10 &92.51&80.13&89.43& 91.91 & \textbf{93.72}**\\
\multirow{8}{*}{} & \multirow{4}{*}{} & 20\% &83.51& 93.20 & \underline{93.25} & 92.87 & 84.76 &77.61&80.67&89.65& 93.10 & \textbf{94.42}**\\
\multirow{8}{*}{} & \multirow{4}{*}{} & 40\% &83.46& 93.43 & 93.25 &82.89&83.13& 92.95 & 85.85 &90.13& \underline{93.69} & \textbf{94.56}**\\
\multirow{8}{*}{} & \multirow{4}{*}{} & 60\% &83.42& 93.61 & 93.34 &85.02&83.82& 93.15 & 86.49&90.62& \underline{93.80} & \textbf{94.87}**\\
\multirow{8}{*}{} & \multirow{4}{*}{} & 80\% &83.32& 93.69 & 93.57 &84.95&84.06& 93.31 & 86.64 &92.15&\underline{94.34} & \textbf{95.03}**\\
\hline
\multirow{8}{*}{YELP} & \multirow{5}{*}{Macro-F1} & 10\% &54.03& 76.85 & 86.86 &53.96&54.04&	72.42 & 73.78&60.18 & \underline{90.96} & \textbf{91.48}\\
\multirow{8}{*}{} & \multirow{4}{*}{} & 20\% &54.07& 77.24 & 87.86 &53.96&54.07&	75.06 & 74.01&60.59 & \underline{91.57} & \textbf{92.04}\\
\multirow{8}{*}{} & \multirow{4}{*}{} & 40\% &54.07& 78.48 & 89.85 &54.00&54.07& 76.49 & 77.09 &66.08& \underline{92.84} & \textbf{92.91} \\
\multirow{8}{*}{} & \multirow{4}{*}{} & 60\% &54.00& 78.58 & 90.58 &53.96&54.00& 77.09 & 80.03 &67.44& \underline{93.03} & \textbf{93.43} \\
\multirow{8}{*}{} & \multirow{4}{*}{} & 80\% &53.82& 78.93 & 90.57 &53.70&53.82& 77.93 & 81.18 &68.73& \underline{93.59} & \textbf{93.74}*\\
\cline{2-13}
\multirow{8}{*}{} & \multirow{5}{*}{Micro-F1} & 10\% &73.01& 75.98 & 86.68 &72.86&73.03& 78.52 & 77.36 &74.83& \underline{90.29} & \textbf{91.01}\\
\multirow{8}{*}{} & \multirow{4}{*}{} & 20\% &73.06& 78.85 & 87.84 &72.89&73.06& 79.88 & 78.55 &75.06& \underline{90.87} & \textbf{91.74}** \\
\multirow{8}{*}{} & \multirow{4}{*}{} & 40\% &73.14& 79.92 & 89.86 &72.95&73.14& 80.68 & 80.48&76.77 & \underline{92.19} & \textbf{92.58} \\
\multirow{8}{*}{} & \multirow{4}{*}{} & 60\% &72.97& 79.97 & 90.64 &72.97&72.97& 81.00 & 82.62 &77.46& \underline{92.42} & \textbf{92.97}*\\
\multirow{8}{*}{} & \multirow{4}{*}{} & 80\% &72.82& 80.41 & 90.62 &72.78&72.82& 81.55 & 83.55 &78.04& \underline{93.03} & \textbf{93.35}*\\
\hline
\end{tabular}
}
\label{table3}
\end{table*}

\subsection{Classification Results}
We first evaluate the performance of \name\ on the node classification task,
where we use Macro-F1 and Micro-F1 as metrics.
For both of them, 
the larger the value,
the better the model performance.
For semi-supervised methods,
labeled nodes are divided into training, validation, and test sets in the ratio of 10\%, 10\%, and 80\%, respectively.
To ensure a fair comparison between semi-supervised and unsupervised models, 
following~\cite{jin2021heterogeneous},
we only report the classification results on the test set.
For baselines that cannot handle missing attributes,
we complete missing attributes by 
averaging attributes from a node's neighbors. 
For datasets ACM, DBLP and YELP, we use learned node embeddings to further train a linear SVM classifier \cite{fu2020magnn,jin2021heterogeneous} with different training ratios from 10\% to 80\%. For the largest dataset AMiner, following~\cite{tian2022heterogeneous}, we select 20, 40 and 60 labeled nodes per class as training set, respectively and further train a Logistic Regression model. 
We report the average Macro-F1 and Micro-F1 results over 10 runs to evaluate the model. 

\subsubsection{With missing attributes.}
We first show the 
classification results 
in the presence of missing features.  
Since the
results of most baselines on these benchmark datasets are public, we directly report these results from their original papers. For cases where the results are missing, we obtain them from~\cite{he2022analyzing,tian2022heterogeneous}.
The results are shown in Table~\ref{table3} and Table~\ref{table4}. 
From the tables,
we observe that
(1) HGCA and \name,
which are specially designed to handle  HINs with missing attributes,
generally perform better than other baselines.
(2) 
While HGCA can perform well in some cases,
it fails to run on the  
AMiner dataset because
it explicitly requires the existence of some attributed nodes in the graph.
For the non-attributed dataset,
it cannot be applied.
Further,
HGCA generates features in the raw high-dimensional space for non-attributed nodes,
while \name\ constructs low-dimensional attributes for them.
The former is more likely to contain noise,
which adversely affects the model performance.
(3)
\name\ achieves the best results in 30 out of 36 cases,
where 60\% results are statistically significant.
This also shows the effectiveness of our method.


\subsubsection{With inaccurate attributes.}
To verify the robustness of the model when tackling inaccurate attributes,
following~\cite{chen2023uncertainty},
we corrupt raw node features with random Gaussian noise $\mathcal{N}(0,\sigma)$, 
where $\sigma$ is computed by the standard deviation of the bag-of-words representations of all the nodes in each graph. 
In particular,
we compare \name\ with 
HGMAE~\cite{tian2022heterogeneous} and HGCA~\cite{he2022analyzing} because they are all self-supervised models that are specially designed for HINs.
We vary the noise level and 
the results are given in Table~\ref{table:noise}.
Note that since {all the nodes in AMiner are non-attributed},
we cannot corrupt node attributes and thus exclude the dataset.
From the table,
we observe that: 
(1) With the increase of the noise level,
the performance of all the three methods drops,
but
\name\ is more robust against HGMAE and HGCA. 
\name\ constructs low-dimensional features for non-attributed nodes and it can rectify inaccurate attributes for attributed nodes with feature reconstruction.
(2)
\textbf{As the noise increases,
the advantages of \name\ over others are more statistically significant (see results on ACM and YELP)}.
This further 
demonstrates that \name\ can well deal with inaccurate attributes in HINs.

\begin{table}[!hbpt]
\centering
\caption{Node classification results (\%) on AMiner.
}
\label{table4}
\resizebox{0.95\linewidth}{!}{
\begin{tabular}{c|c|cccccc}
\hline
Metric & Split &Mp2vec&DGI& DMGI & HGMAE&HGCA& \name\ \\
\hline
\multirow{3}{*}{Macro-F1} &20&60.82&62.39&63.93&\textbf{72.28}&&\underline{69.20}\\
\multirow{3}{*}{} &40&69.66&63.87&63.60&\underline{75.27}&-&\textbf{75.76} \\
\multirow{3}{*}{} &60&63.92&63.10&62.51&\underline{74.67}&&\textbf{75.31}* \\
\cline{1-8}
\multirow{3}{*}{Micro-F1} &20&54.78&51.61&59.50&\textbf{80.30}&&\underline{77.09}\\
\multirow{3}{*}{} &40&64.77&54.72&61.92&\underline{82.35}&-&\textbf{82.56} \\
\multirow{3}{*}{} &60&60.65&55.45&61.15&\underline{81.69}&&\textbf{82.14}* \\
\cline{2-8}
\hline
\end{tabular}}
\label{table4}
\end{table}

\begin{table}[!hbpt]
\centering
\caption{Node classification results (\%) on MAGNN.}
\label{table:quality}
\resizebox{0.95\linewidth}{!}
{
\begin{tabular}{c||c|c|cccc}
\hline
\multirow{2}{*}{Dataset} & \multirow{2}{*}{Metric} & Train & MAGNN  & MAGNN & MAGNN & MAGNN\\
\multirow{2}{*}{} & \multirow{2}{*}{} & -ing & -AVG  & -onehot & -AC & -\name\\
\hline
\multirow{10}{*}{ACM} & \multirow{5}{*}{} & 10\% & 88.82&89.69	&{92.92}&\textbf{93.34}*\\
\multirow{10}{*}{} & Macro & 20\% & 89.41&	90.61&	{93.34}	&\textbf{94.98}**\\
\multirow{10}{*}{} & -F1 & 40\% & 89.83&	92.48&	{93.72}&	\textbf{94.35}**\\
\multirow{10}{*}{} & \multirow{5}{*}{} & 60\% & 90.18&	93.12&	{94.01}&	\textbf{94.98}*
 \\
\multirow{10}{*}{} & \multirow{5}{*}{} & 80\% &90.11	&93.20&	{94.08}&	\textbf{94.76}*
 \\
\cline{2-7}
\multirow{10}{*}{} & \multirow{5}{*}{} & 10\% & 88.81&89.92	&{92.33}&\textbf{93.52}**\\
\multirow{10}{*}{} & Micro & 20\% & 89.36	&90.58	&{93.21}&	\textbf{94.16}**
 \\
\multirow{10}{*}{} & -F1 & 40\% & 89.81&	92.93&	{93.60}&	\textbf{94.49}**
 \\
\multirow{10}{*}{} & \multirow{5}{*}{} & 60\% & 90.11&	93.52&	{93.87}&	\textbf{95.13}**
 \\
\multirow{10}{*}{} & \multirow{5}{*}{} & 80\% & 90.06	&93.65&{93.93}&	\textbf{94.93}**
 \\
\hline
\end{tabular}
}
\end{table}

\begin{table*}[t]
\centering
\caption{Node classification results (\%) with noise in node features. ACM with $\mathbf{\sigma=0.0322}$, YELP with $\mathbf{\sigma=0.1860}$, DBLP with $\mathbf{\sigma=0.0386}$. We highlight the best score on each dataset in bold, where ** and * denote p-value $<0.01$ and $<0.05$, respectively.}
\label{table:noise}
\resizebox{0.8\linewidth}{!}
{
\begin{tabular}{c|c||c|ccc||ccc||ccc}
\hline
\rule{0pt}{8pt}
\multirow{3}{*}{ Noise}&\multirow{3}{*}{Metric} & \multirow{3}{*}{Training}&\multicolumn{9}{c}{Datasets}\\
\cline{4-12}
\rule{0pt}{8pt}
\multirow{3}{*}{}&\multirow{3}{*}{} & \multirow{3}{*}{}& \multicolumn{3}{c||}{ACM}& \multicolumn{3}{c||}{DBLP}& \multicolumn{3}{c}{YELP}\\
\cline{4-12}
\rule{0pt}{8pt}
\multirow{3}{*}{}&\multirow{3}{*}{} & \multirow{3}{*}{}& HGMAE  & HGCA & \name\ & HGMAE  & HGCA & \name\ & HGMAE  & HGCA & \name\\
\hline
\multirow{10}{*}{$\mathcal{N}(0,\sigma)$} & \multirow{5}{*}{Macro-F1} & 10\% &84.32&\textbf{91.46}&91.15&88.47&{91.12}&\textbf{93.93}**&62.13&90.23&\textbf{90.81}*\\
\multirow{10}{*}{} & \multirow{5}{*}{} & 20\% &84.44&91.62&\textbf{91.91}&88.69&91.30&\textbf{94.22}**&62.59&91.09&\textbf{91.14}\\
\multirow{10}{*}{} & \multirow{5}{*}{} & 40\% &85.94&91.92&\textbf{92.02}&89.30&91.72&\textbf{94.39}**&66.03&\textbf{91.69}&91.30\\
\multirow{10}{*}{} & \multirow{5}{*}{} & 60\% &86.59&{92.15}&\textbf{92.16}&89.41&92.01&\textbf{94.51}** &67.26&\textbf{92.05}&91.10 \\
\multirow{10}{*}{} & \multirow{5}{*}{} & 80\% &87.14&92.14&\textbf{92.22} &90.27&92.10&\textbf{93.98}**&66.65&91.51&\textbf{91.60}\\
\cline{2-12}
\multirow{10}{*}{} & \multirow{5}{*}{Micro-F1} & 10\% &84.67&\textbf{91.47}&91.10&89.12&92.27&\textbf{94.39}**&68.04&90.01&\textbf{90.45}*\\
\multirow{10}{*}{} & \multirow{5}{*}{} & 20\% &84.90&91.63&\textbf{91.84}&89.34&92.42&\textbf{94.64}**&72.05&{90.79}&\textbf{90.80}\\
\multirow{10}{*}{} & \multirow{5}{*}{} & 40\% &86.29&91.94&\textbf{91.96}&89.53&92.75&\textbf{94.83}**&73.21&\textbf{91.32}&90.95\\
\multirow{10}{*}{} & \multirow{5}{*}{} & 60\% &86.82&\textbf{92.16}&92.14&90.06&93.01&\textbf{94.90}**&75.95&\textbf{91.67}&90.77\\
\multirow{10}{*}{} & \multirow{5}{*}{} & 80\% &87.41&92.22&\textbf{92.27}&91.29&93.07&\textbf{94.45}**&75.28&91.24&\textbf{91.34}\\
\hline
\multirow{10}{*}{$\mathcal{N}(0,2\sigma)$} & \multirow{5}{*}{Macro-F1} & 10\% &82.43&90.77&\textbf{91.28}**&87.69&88.64&\textbf{93.54}**&60.79&89.87&\textbf{90.18}*\\
\multirow{10}{*}{} & \multirow{5}{*}{} & 20\% &82.54&91.33&\textbf{91.70}*&87.81&89.79&\textbf{93.90}**&61.53&90.63&\textbf{90.68}\\
\multirow{10}{*}{} & \multirow{5}{*}{} & 40\% &84.57&91.77&\textbf{91.82}&88.36&90.30&\textbf{93.92}**&62.39&\textbf{91.01}&90.98\\
\multirow{10}{*}{} & \multirow{5}{*}{} & 60\% &85.29&{92.06}&\textbf{92.17}&88.86&91.00&\textbf{93.99}**&64.29&\textbf{91.27}&90.79\\
\multirow{10}{*}{} & \multirow{5}{*}{} & 80\% &85.51&{92.22}&\textbf{92.46}&89.07&90.87&\textbf{93.34}**&63.33&90.80&\textbf{91.36}**\\
\cline{2-12}
\multirow{10}{*}{} & \multirow{5}{*}{Micro-F1} & 10\% &82.95&{90.72}&\textbf{91.22}**&88.15&89.12&\textbf{94.00}**&67.96&{89.44}&\textbf{89.66} \\
\multirow{10}{*}{} & \multirow{5}{*}{} & 20\% &83.32&{91.21}&\textbf{91.62}*&88.91&90.28&\textbf{94.30}**&71.38&\textbf{90.17}&{90.08}\\
\multirow{10}{*}{} & \multirow{5}{*}{} & 40\% 
&85.14&\textbf{91.75}&\textbf{91.75}&89.37&90.69&\textbf{94.34}** &72.61&{90.46}&\textbf{90.50}\\
\multirow{10}{*}{} & \multirow{5}{*}{} & 60\% &85.86&{92.03}&\textbf{92.11}&89.84&91.36&\textbf{94.35}**&73.17&\textbf{90.72}&{90.24}\\
\multirow{10}{*}{} & \multirow{5}{*}{} & 80\%
&86.18&{92.27}& \textbf{92.45}&90.01&91.17&\textbf{93.83}**&73.02&{90.72}&\textbf{90.84} \\
\hline
\multirow{10}{*}{$\mathcal{N}(0,10\sigma)$} & \multirow{5}{*}{Macro-F1} & 10\% &77.92&{85.49}&\textbf{86.51}**&87.47&88.36&\textbf{92.89}**&60.28&{75.09}&\textbf{76.34}**\\
\multirow{10}{*}{} & \multirow{5}{*}{} &20\%
&78.08&{85.68}&\textbf{87.15}**&87.56&89.17&\textbf{93.06}**&62.06&{77.06}&\textbf{77.98}**\\
\multirow{10}{*}{} & \multirow{5}{*}{} & 40\% &79.31&{85.92}&\textbf{89.19}**&88.12&89.87& \textbf{93.02}**&62.24&{77.79}&\textbf{78.59}** \\
\multirow{10}{*}{} & \multirow{5}{*}{} & 60\% &80.05&{86.06}&\textbf{87.28}**&88.48&90.46&\textbf{93.51}**&62.78&{78.23}&\textbf{78.46}\\
\multirow{10}{*}{} & \multirow{5}{*}{} & 80\% &79.87&{85.80}&\textbf{88.17}**&88.80&90.54&\textbf{93.25}**&61.26&{78.07}&\textbf{78.63}*
 \\
\cline{2-12}
\multirow{10}{*}{} & \multirow{5}{*}{Micro-F1} & 10\% &78.47&{85.27}&\textbf{86.43}**&88.01&88.57&\textbf{93.38}**&64.57&{78.87}&\textbf{79.55}**
 \\
\multirow{10}{*}{} & \multirow{5}{*}{} & 20\% &78.72&{85.46}&\textbf{87.08}**&88.36&89.44&\textbf{93.52}**&67.67&{80.51}&\textbf{80.93}*
 \\
\multirow{10}{*}{} & \multirow{5}{*}{} & 40\% &79.85&{85.72}&\textbf{89.12}**&88.82&90.07&\textbf{93.50}**&71.88&{81.03}&\textbf{81.32}\\
\multirow{10}{*}{} & \multirow{5}{*}{} & 60\% &80.58&{85.87}&\textbf{87.28}**&89.20&90.06&\textbf{93.94}**&73.47&\textbf{81.29}&{81.19}\\
\multirow{10}{*}{} & \multirow{5}{*}{} & 80\% &80.45&{85.71}&\textbf{88.17}**&89.51&91.12&\textbf{93.77}**&73.31&{81.02}&\textbf{81.48}*\\
\hline

\end{tabular}
}
\end{table*}

\subsection{Quality of Generated Attributes}
\label{sec:extension}
To evaluate the quality of generated attributes by \name, 
we take the raw graph with generated/reconstructed attributes for both non-attributed and attributed nodes
as input, which is further fed
into 
the MAGNN~\cite{fu2020magnn} model for node classification.
We call the method MAGNN-\name.
We compare it with three variants that have different attribute completion strategies:
MAGNN-AVG,
MAGNN-onehot and
MAGNN-AC~\cite{jin2021heterogeneous}.
For a non-attributed node,
they complete the missing attributes
by averaging its neighboring attributes, 
defining one-hot encoded vectors, and calculating the  weighted average of neighboring attributes with the attention mechanism, respectively.
For MAGNN-\name,
we generate low-dimensional attributes for non-attributed nodes and replace raw attributes with the reconstructed high-dimensional ones for attributed nodes.
The results are given in Table~\ref{table:quality}.
Due to the space limitation, we choose ACM as the representative dataset and the full results are given in {Appendix~\ref{sec:app:quality}.} 
We find that
MAGNN-\name\ achieves the best performance that are also statistically significant.
On the on hand, \name\ can effectively utilize fine-grained semantic information from both node-level and attribute-level embeddings to generate high-quality attributes.
On the other hand, \name\ can denoise the original high-dimensional inaccurate attributes. 
This ensures the effectiveness of the generated attributes by \name.

\subsection{Link Prediction }
\label{section:APPD}
We further evaluate the performance of \name\ on the link prediction task, 
where we treat the connected nodes in the HIN as positive node pairs and other unconnected nodes as negative node pairs. 
For each edge type,
we divide the positive node pairs into 85\% training set, 5\% validation set and 10\% testing set.
We also randomly add the same number of negative node pairs to the dataset. 
We use AUC (Area Under the Curve) and AP (Average Precision) as the evaluation metrics. 
We fine-tune hyper-parameters with a patience of 100 by validation set. 
For all methods, 
we run experiments 5 times 
to report the mean and standard deviation values,
as shown in Table~\ref{table:lp}.
From the table, we observe that:
(1)
\name\ outperforms baselines across various relationships in the majority of cases.
(2) While SeeGera and SIG-VAE perform well on the P-A relation of ACM dataset,
they are designed for homogeneous graphs that disregard edge types and cannot provide robust performance.
All these results show the superiority of \name\ on heterogeneous graphs.

\begin{table*}[t]
\centering
\caption{Link prediction results (\%) in the form of mean ± std. 
We highlight the best score in bold and underline the runner-up.}
\renewcommand{\arraystretch}{1.2}
\resizebox{0.9\linewidth}{!}
{
\begin{tabular}{c||cc|cccccccc}
\hline
Metric & Dataset & Relation & VGAE & SIG-VAE  & CAN & RGCN &HGT& SeeGera & MHGCN & \name\\
\hline
\multirow{8}{*}{AUC} & \multirow{2}{*}{ACM} & P-A & $90.87\pm 1.42$ & $\underline{92.35\pm 0.79}$ & $91.47\pm 0.26$ & $78.29\pm 0.72$& $77.83\pm 0.11$ & $\textbf{92.86}\pm \textbf{0.41}$&$88.52\pm 0.57$ & $90.32 \pm 0.39$\\
\multirow{8}{*}{} & \multirow{2}{*}{} & P-S & $90.96\pm 2.21$ & $89.95\pm 0.46$  & $\underline{95.35\pm 0.24}$ & $86.07\pm 4.94$& $91.40\pm 2.91$& $90.23\pm 0.54$&$89.34\pm 0.64$ & $\textbf{96.10}\pm \textbf{ 0.35}$**\\
\cline{2-11}
\multirow{8}{*}{} & \multirow{3}{*}{DBLP} & P-A & $\underline{91.61\pm 0.61}$ & $89.34\pm 0.30$  & $89.50 \pm 0.41$& $76.48\pm 0.98$& $85.02\pm 2.24$ & $89.10 \pm 1.27$& $87.43 \pm 0.08$& $\textbf{92.76}\pm \textbf{0.38}$**\\
\multirow{8}{*}{} & \multirow{3}{*}{} & P-T & $\underline{91.60\pm 0.33}$ & $89.89\pm 0.20$  & $91.26\pm 0.09$ & $81.33\pm 1.36$& $88.71\pm 1.17$& $90.87\pm 0.16$& $88.55 \pm 0.15$ &$\textbf{92.58}\pm \textbf{0.22}$**\\
\multirow{8}{*}{} & \multirow{3}{*}{} & P-V & $92.09\pm 0.49$ & $89.15\pm 0.13$  & $94.21\pm 0.32$ & $71.66\pm 2.59$&  $\underline{96.11\pm 1.43}$& $88.74\pm 0.39$&$94.94 \pm 0.09$ & $\textbf{96.35}\pm \textbf{0.18}$\\
\cline{2-11}
\multirow{8}{*}{} & \multirow{3}{*}{YELP} & B-U & $90.65\pm 0.14$ & $88.96\pm 0.95$  & $90.40\pm 0.72$ & $91.62\pm 0.41$& $\underline{92.54\pm 0.21}$& $91.05\pm 1.06$&$90.78\pm 1.23$ & $\textbf{93.61}\pm \textbf{0.14}$**\\
\multirow{8}{*}{} & \multirow{3}{*}{} & B-S & $84.52\pm 0.74$ & $86.45\pm 0.26$  & $92.18\pm 0.08$ & $78.27\pm 1.92$&  $\underline{97.66}\pm \underline{0.80}$& $89.23\pm 0.72$& $92.01\pm 1.56$ &$\textbf{97.78}\pm \textbf{0.22}$\\
\multirow{8}{*}{} & \multirow{3}{*}{} & B-L & $86.18\pm 0.59$  & $85.89\pm 0.51$ & $86.46\pm 0.26$ & $84.83\pm 0.28$&  $\underline{90.80\pm 1.83}$& $85.91\pm 0.05$& $91.77\pm 1.45$& $\textbf{91.43}\pm \textbf{0.16}$\\
\hline
\multirow{8}{*}{AP} & \multirow{2}{*}{ACM} & P-A & $\underline{92.40\pm 0.44}$ & $\textbf{93.23}\pm \textbf{0.71}$  & $91.93\pm 0.24$ & $72.81\pm 0.61$& $73.27\pm 0.93$& $92.12\pm 0.40$& $88.94\pm 0.53$& $91.06 \pm 0.42$\\
\multirow{8}{*}{} & \multirow{2}{*}{} & P-S & $91.95\pm 0.21$ & $91.38\pm 0.92$  & $\underline{93.28\pm 0.39}$ & $86.47\pm 5.34$& $87.80\pm 3.95$& $91.64\pm 0.47$& $90.21\pm 0.51$& $\textbf{95.65}\pm \textbf{0.22}$**\\
\cline{2-11}
\multirow{8}{*}{} & \multirow{3}{*}{DBLP} & P-A & $\underline{91.20\pm 0.85}$ & $87.87\pm 0.53$  & $87.83\pm 0.55$ & $66.93\pm 0.51$& $80.64\pm 0.82$& $88.37\pm 0.73$& $89.95 \pm 0.06$ &$\textbf{92.34}\pm \textbf{0.48}$**\\
\multirow{8}{*}{} & \multirow{3}{*}{} & P-T & $\underline{92.37\pm 0.29}$ & $91.75\pm 0.19$  & $91.54\pm 0.07$ & $85.02\pm 0.59$& $87.31\pm 1.73$& $90.96\pm 0.06$& $90.39 \pm 0.11$ &$\textbf{92.92}\pm \textbf{0.18}$**\\
\multirow{8}{*}{} & \multirow{3}{*}{} & P-V & $93.62\pm 0.55$ & $91.14\pm 0.10$  & $94.91\pm 0.29$& $60.51\pm 0.11$& $\underline{95.09\pm 1.72}$ & $90.97\pm 0.37$& $93.48 \pm 0.14$ &$\textbf{95.70}\pm \textbf{0.17}$\\
\cline{2-11}
\multirow{8}{*}{} & \multirow{3}{*}{YELP} & B-U & $89.97\pm 0.38$ & $88.02\pm 0.77$  & $88.64\pm 0.52$ & $\underline{91.05\pm 0.37}$& $90.71\pm 0.32$& $90.09\pm 1.64$&$89.63\pm 1.29$ & $\textbf{92.65}\pm \textbf{0.04}$**\\
\multirow{8}{*}{} & \multirow{3}{*}{} & B-S & $83.85\pm 1.81$ & $88.29\pm 0.69$ & $93.32\pm 0.80$ & $81.28\pm 0.49$& $\underline{97.34\pm 1.41}$ & $91.72\pm 0.41$&$91.82\pm 1.68$ & $\textbf{97.52}\pm \textbf{0.28}$\\
\multirow{8}{*}{} & \multirow{3}{*}{} & B-L & $85.75\pm 0.19$ & $83.21\pm 0.17$  & $\underline{86.84\pm 0.61}$ & $81.28\pm 0.48$&$85.85\pm 2.16$& $83.50\pm 0.77$& $90.93\pm 1.32$& $\textbf{88.58}\pm \textbf{0.97}$**\\
\hline
\end{tabular}
}
\label{table:lp}
\end{table*}

\subsection{Efficiency Study}
\begin{figure}[htbp]
    \centering
    \includegraphics[width=0.5\linewidth]{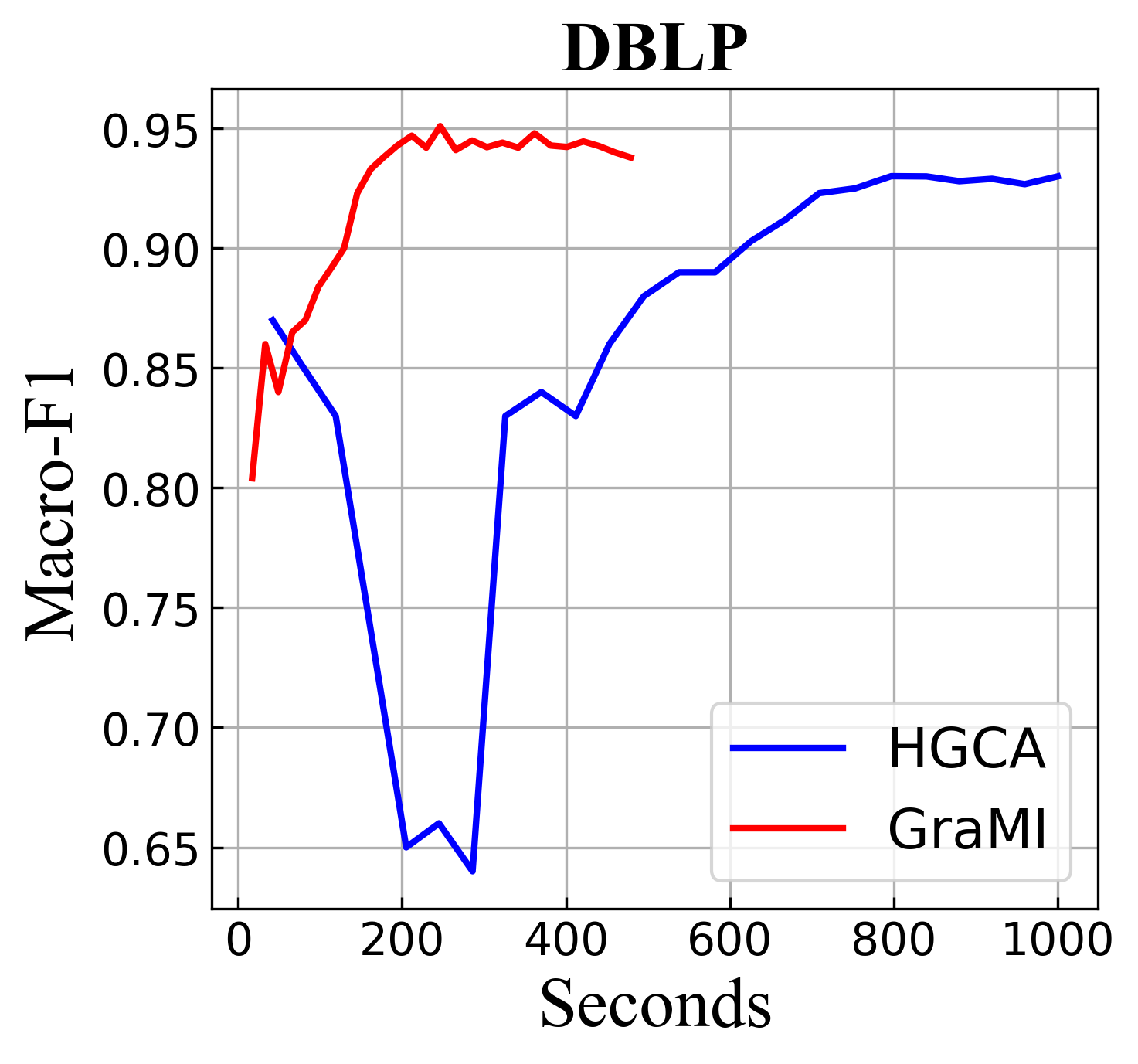}
    \caption{Efficiency study on DBLP.}
    \label{fig:speed}
\end{figure}

We next evaluate the efficiency of \name.  
To ensure a fair comparison, we measure the training time of both \name\ and HGCA~\cite{he2022analyzing}, because they rank as the top two best performers. It's noteworthy that both models are self-supervised, which are specially designed for HINs with missing attributes.
Further, since HGCA cannot be applied on the largest AMiner dataset,
we take the second largest dataset DBLP as an example. 
For illustration, 
we utilize 80\% of labeled nodes as our training set. 
From the figure~\ref{fig:speed},
we see that 
\name\ converges fast and 
demonstrates consistent performance growth. On DBLP,
\name\ achieves almost $4\times$ speedup than HGCA. 
This further shows that \name\ is not only effective but efficient.

\begin{figure}[!htbp]
    \centering
    \includegraphics[width=0.9\linewidth]{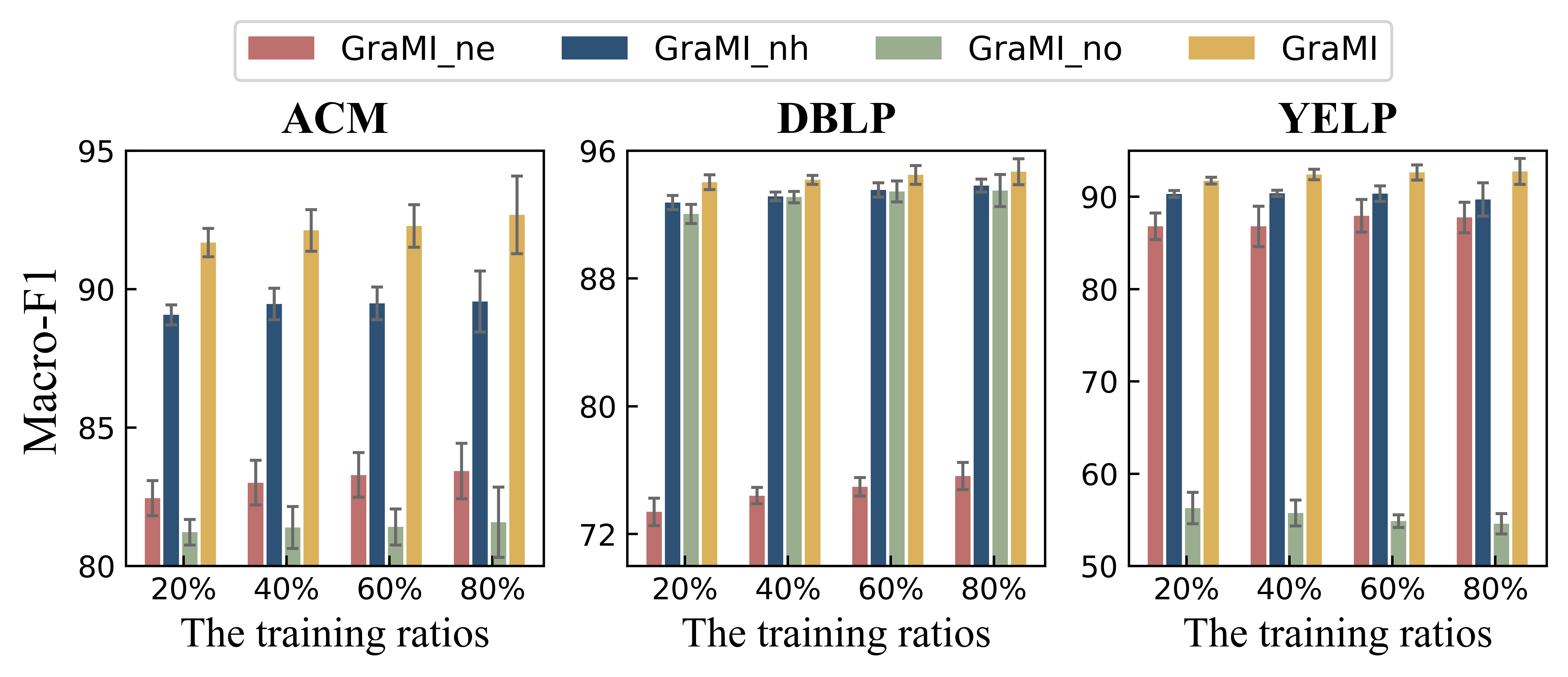}
    \caption{The ablation study results.}
    \label{fig:ablation}
\end{figure} 

\subsection{Ablation Study}
We next conduct an ablation study to comprehend three main components of \name.
To show the importance of edge reconstruction, we remove the term $\mathcal{L}_{edge}$ in Equation~\ref{L1} and call the variant \textbf{\name\_ne} (\textbf{n}o \textbf{e}dge reconstruction).
Similarly, to understand the importance of attribute reconstruction, 
we remove the term $\mathcal{L}_{attr}$ and $\mathcal{L}_{rmse}$ respectively.
We call the two variants 
\textbf{\name\_nh} (\textbf{n}o \textbf{h}idden embedding reconstruction)
and \textbf{\name\_no} (\textbf{n}o \textbf{o}riginal attribute reconstruction).
Finally, we compare \name\ with them on three benchmark datasets with different training ratios and show the results in Figure~\ref{fig:ablation}. We can see that
(1) \name\ consistently outperforms all variants across the three datasets.
(2) The performance gap between \name\ and {\name\_ne} ({\name\_no}) shows the importance of edge reconstruction (raw attributes reconstruction) in learning node embeddings.
(3) \name\ performs better than {\name\_nh}, which shows that the hidden embedding reconstruction can help generate better embeddings for both non-attributed nodes and attributed nodes, and further boost the model performance.

\subsection{Hyper-parameter Sensitivity Analysis}
\begin{figure}[!htbp]
    \centering
    \includegraphics[width=0.9\linewidth]{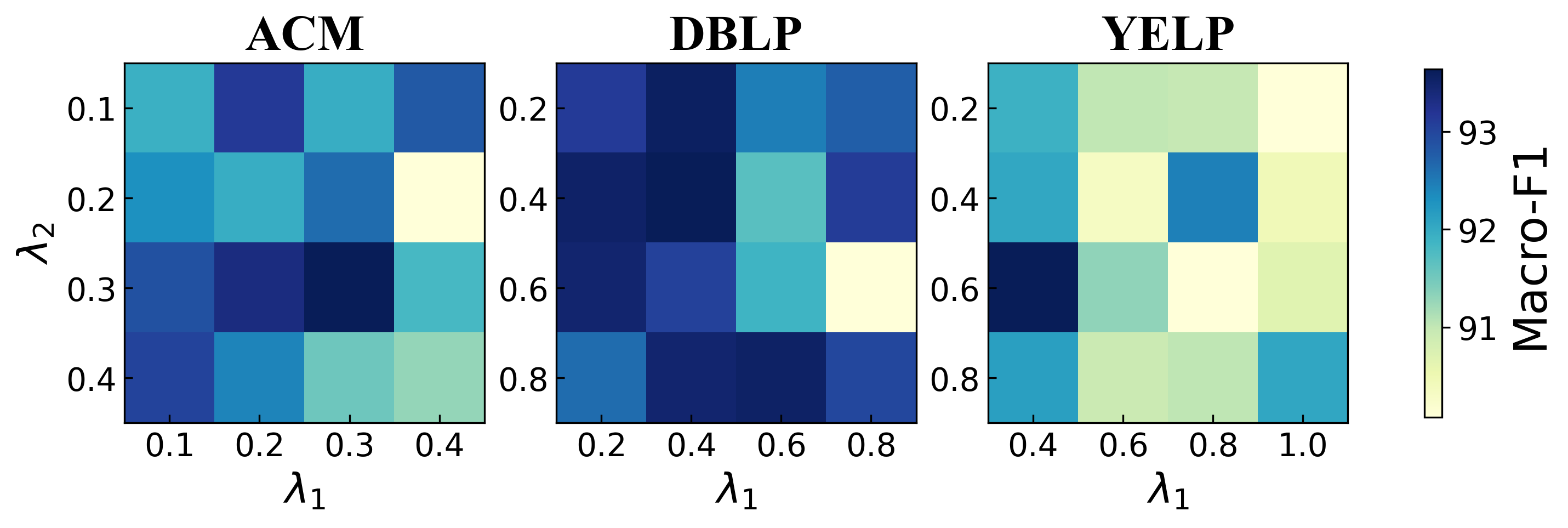}
    \caption{Hyper-parameter sensitivity analysis.}
    \label{fig:dblp_yelp}
\end{figure}

We end this section with a sensitivity analysis on the hyper-parameters of \name,
i.e., two coefficients $\lambda_1$ and $\lambda_2$, which control the importance of hidden and raw feature reconstruction, respectively.
In our experiments, we vary one hyper-parameter each time with the other fixed.
Figure~\ref{fig:dblp_yelp} shows the Macro-F1 scores for \name\ on three datasets.
For Micro-F1 scores,
we observe similar results, which are thus omitted due to the space limitation.
From the figure,  we see that for both hyper-parameters, \name\ can give stable
performances over a wide range of values.
This demonstrates the insensitivity of \name\ on hyper-parameters. 

\section{CONCLUSION}
In this paper, we studied generative SSL on heterogeneous graph and proposed \name.
Specifically, \name\ employs hierarchical variational inference
to generate both node- and attribute-level embeddings.
Importantly, 
\name\ expands attribute inference in the low-dimensional hidden space to address the problems of missing and inaccurate attributes in HINs.
By jointly generating node-level and attribute-level embeddings,
fine-grained semantic information can be obtained for generating attributes. Besides,
\name\ reconstructs the low-dimensional node embedding to alleviate the adverse effect
of noise attribute and enhance
the feature information of missing attribute.
We conducted extensive experiments to evaluate the performance of \name.
The results shows that \name\ is very effective to deal with above problems and leads to superior performance.
\section{Acknowledgments}
This work is supported by Shanghai Science and Technology Committee General Program No. 22ZR1419900.

\bibliographystyle{ACM-Reference-Format}
\bibliography{sample-sigconf}

\clearpage

\appendix


\section{Notations}
We summarize key notations and their explanations used in the paper as shown in Table~\ref{notations}.
\begin{table}[!htbp]
\centering
\caption{Notations and explanations}
{
\begin{tabular}{c|c}
\hline
Notations & explanations\\
\hline
$\mathcal{V}$ & The set of nodes\\
$\mathcal{E}$ & The set of edges\\
$\mathcal{A}$ & The set of node attributes\\
$\mathcal{T}$ & The set of node types\\
$\mathcal{R}$ & The set of edge types\\
$\text{A}$ & Adjacency matrix\\
$\mathcal{T^+}$ & The set of attributed node types\\
$\mathcal{T^-}$ & The set of non-attributed node types\\
$\text{X}$ & Raw node features\\
$\tilde{\text X}$ & Hidden node features\\
$\tilde{\text X}'$ & Reconstructed hidden node features\\
 $\mathrm Z^\mathcal V$ & Node embeddings\\
 $\mathrm Z^\mathcal A$ & Attribute embeddings\\
\hline
\end{tabular}
}
\label{notations}
\end{table}

\section{Datasets}
\begin{itemize}
    \item \textbf{ACM}: This dataset is extracted from the Association for Computing Machinery website (ACM). It contains 4019 papers (P), 7167 authors (A) and 60 subjects (S). The papers are divided into three classes according to the conference they published. In this dataset, only paper's nodes have original attributes which are bag-of-words representations of their keywords, other nodes have no attribute.
    \item \textbf{DBLP}: This dataset is extracted from the Association for Computing Machinery website (DBLP). It contains 4057 authors (A), 14328 papers (P), 8789 terms (T) and 20 venues (V). Authors are divided into four research areas according to the conferences they submitted. In this dataset, only paper's nodes have directly original attributes which are bag-of-words representations of their keywords, other nodes have no attribute.
    \item \textbf{YELP}: This dataset is extracted from the Yelp Open dataset. It contains 2614 bussinesses (B), 1286 users (U), 4 services (S) and  9 rating levels (L). The business are divided into three classes according to their categories. In this dataset, only bussiness's node have original attributes which are representations about their descriptions, other nodes have no attribute.
    \item \textbf{AMiner}: This dataset is extracted from the AMiner citation network. It contains 6564 papers (P), 13329 authors (A) and 35890 reference (R). The papers are divided into four classes. In this dataset, all types of nodes have no attribute.
\end{itemize}
\begin{table}[!htbp]
    \centering
    \caption{Statistics of datasets.}
    \label{table1}
    \renewcommand{\arraystretch}{1.25}
    \begin{tabular}{c||c|c|c}
    \hline
         Datasets & Nodes & Edges & Attributes\\
         \hline
         ACM & \makecell[l]{paper(P):4019\\author(A):7167\\subject(S):60} &\makecell[l]{P-P:9615\\P-A:13407\\P-S:4019} &\makecell[l]{P:raw\\A:handcrafted\\S:handcrafted}\\
         \hline
         DBLP & \makecell[l]{author(A):4057\\paper(P):14328\\term(T):7723\\venue(V):20} & \makecell[l]{P-A:19645\\P-T:85810\\P-V:14328} & \makecell[l]{A:handcrafted\\P:raw\\T:handcrafted\\V:handcrafted}\\
         \hline
         YELP & {\makecell[l]{business(B):2614 \\ user(U):1286 \\ service(S):4 \\ level(L):9}} & {\makecell[l]{B-U:30838\\B-S:2614\\B-L:2614}} & {\makecell[l]{B:raw\\U:handcrafted\\S:handcrafted\\L:handcrafted}}\\
         \hline
         AMiner & \makecell[l]{paper(P): 6564\\author(A): 13329\\reference(R): 35890} &\makecell[l]{P-A: 18007\\P-R: 58831} &\makecell[l]{P: handcrafted\\A: handcrafted\\S: handcrafted}\\
         \hline
    \end{tabular}
    \label{table1}
\end{table}

\section{Implementation Details} 
\label{section:APPC}
We have implemented \name\ using PyTorch. The model is initialized with Xavier initialization~\cite{glorot2010understanding} and trained using the Adam optimizer~\cite{kingma2014adam}.
Following~\cite{hasanzadeh2019semi},
we set noise distribution
$q_1(\epsilon)$ and $q_2(\epsilon)$
as standard Gaussian distribution.
For homogeneous methods, we treat all nodes and edges as the same type. We fine-tune the hyper-parameters for all methods we have compared to report their best results. For \name, we use the two-layer simple HGNN mentioned above as the backbone for the encoder. For other hyperparameters, we conduct a grid search. The learning rate is adjusted within \{0.001, 0.005, 0.01, 0.05\}, and the dropout rate is selected from \{0.0, 0.3\}. Additionally, we utilize multi-head attention with the number of attention heads K from \{1, 2, 4, 8\}. The embedding dimension is searched from the range \{32, 64, 128, 256\}. For hyperparameters $\lambda_1$ and $\lambda_2$, we select values from the range [0, 1] with a step of 0.1. The number of HGNN layers we use for decoding is selected from \{0, 1, 2\}. 
For fairness, we run all the experiments on a server with 32G memory and a single Tesla V100 GPU. 

\setcounter{equation}{0}
\renewcommand\theequation{A.\arabic{equation}}
\section{Variational Lower Bound}
According to SIVI~\cite{yin2018semi}, the adjacency matrix $\mathrm A$ and the attribute matrix $\mathrm X$ are observed of the heterogenous graph, in order to approximate the true posterior distribution $p(\mathrm Z^\mathcal V,\mathrm Z^\mathcal A|\mathrm A,\mathrm X)$, considering the SIVI we need a variational distribution $h_{\phi}(\mathrm Z^\mathcal V,\mathrm Z^\mathcal A)$ with a variational parameter $\psi$ to minimize $\text{KL}(h_{\phi}(\mathrm Z^\mathcal V,\mathrm Z^\mathcal A)||p(\mathrm Z^\mathcal V,\mathrm Z^\mathcal A|\mathrm A,\mathrm X))$, which is equivalent to maximizing the $\text{ELBO}$~\cite{bishop2013variational}:
\begin{equation}
    \text{ELBO}=\mathrm E_{\mathrm Z^\mathcal V,\mathrm Z^\mathcal A\sim h_{\phi}(\mathrm Z^\mathcal V,\mathrm Z^\mathcal A)}[\log \frac{p(\mathrm Z^\mathcal V,\mathrm Z^\mathcal A,\mathrm A,\mathrm X)}{h_{\phi}(\mathrm Z^\mathcal V,\mathrm Z^\mathcal A)}]=\mathcal{L}.\label{elbo}
\end{equation}

Since $\mathrm Z^\mathcal V$ and $\mathrm Z^\mathcal A$ represent node-level and attribute-level respectively, we assume that they are independent and come from different variational distributions:
\begin{equation}
    \begin{aligned}
    \mathrm{Z}^{\mathcal{V}} \sim h_{\phi_1}(\mathrm{Z}^{\mathcal{V}})=\int_{\psi_1} q_1(\mathrm{Z}^{\mathcal{V}} \mid \psi_1) q_{\phi_1}(\psi_1) \mathrm{d} \psi_1, \\
     \mathrm{Z}^{\mathcal{A}} \sim h_{\phi_2}(\mathrm{Z}^{\mathcal{A}})=\int_{\psi_2} q_2(\mathrm{Z}^{\mathcal{A}} \mid \psi_2) q_{\phi_2}(\psi_2) \mathrm{d} \psi_2.
    \end{aligned}
\end{equation}
Thereby, the $h_{\phi}(\mathrm{Z}^\mathcal V,\mathrm{Z}^\mathcal A)$ is a mean-field distribution that can be factorized as:
\begin{equation}
    h_{\phi}(\mathrm{Z}^\mathcal V,\mathrm{Z}^\mathcal A)=h_{\phi_1}(\mathrm{Z}^\mathcal V)h_{\phi_2}(\mathrm{Z}^\mathcal A).\label{meanfield}
\end{equation}
The joint distribution $p(\mathrm{Z}^\mathcal V,\mathrm{Z}^\mathcal A,\mathrm A,\mathrm X)$ can be represented as:
\begin{equation}
    p(\mathrm{Z}^\mathcal V,\mathrm{Z}^\mathcal A,\mathrm A,\mathrm X)=p(\mathrm{Z}^\mathcal V)p(\mathrm{Z}^\mathcal A)p(\mathrm A|\mathrm{Z}^\mathcal V)p(\mathrm X|\mathrm{Z}^\mathcal V,\mathrm{Z}^\mathcal A).\label{joint}
\end{equation}
 Due to the concavity of the logarithmic function, we use Jensen’s inequality to derive a lower bound for the ELBO by substituting~\eqref{meanfield} and~\eqref{joint} into~\eqref{elbo}:
\begin{equation}
  \begin{aligned}
    \mathcal{L} 
      &= \mathrm E_{\mathrm{Z}^\mathcal V\sim{h_{\phi_1}(\mathrm{Z}^\mathcal V),\mathrm{Z}^\mathcal A\sim h_{\phi_2}(\mathrm{Z}^\mathcal A)}}\\ &\left[\log\frac{p(\mathrm{Z}^\mathcal V)p(\mathrm{Z}^\mathcal A)p(\mathrm A|\mathrm{Z}^\mathcal V)p(\mathrm X|\mathrm{Z}^\mathcal V,\mathrm{Z}^\mathcal A)}{h_{\phi_1}(\mathrm{Z}^\mathcal V)h_{\phi_2}(\mathrm{Z}^\mathcal A)}\right] \\
      &\ge \mathrm E_{\mathrm{Z}^\mathcal V\sim{h_{\phi_1}(\mathrm{Z}^\mathcal V)}}\log p(\mathrm A|\mathrm{Z}^\mathcal V)\\&+\mathrm E_{\mathrm{Z}^\mathcal V\sim{h_{\phi_1}(\mathrm{Z}^\mathcal V),\mathrm{Z}^A\sim h_{\phi_2}(\mathrm{Z}^\mathcal A)}}\log p(\mathrm X|\mathrm{Z}^\mathcal V,\mathrm{Z}^\mathcal A)\\& +\mathrm E_{\mathrm{Z}^\mathcal V\sim{h_{\phi_1}(\mathrm{Z}^\mathcal V)}}\left[\log\frac{p(\mathrm{Z}^\mathcal V)}{h_{\phi_1}(\mathrm{Z}^\mathcal V)}\right]\\&+ \mathrm E_{\mathrm{Z}^\mathcal A\sim h_{\phi_2}(\mathrm{Z}^\mathcal A)}\left[\log\frac{p(\mathrm{Z}^\mathcal A)}{h_{\phi_2}(\mathrm{Z}^\mathcal A)}\right]\\
      &=\mathrm E_{\mathrm{Z}^\mathcal V\sim{h_{\phi_1}(\mathrm{Z}^\mathcal V)}}\log p(\mathrm A|\mathrm{Z}^\mathcal V)\\&+\mathrm E_{\mathrm{Z}^\mathcal V\sim{h_{\phi_1}(\mathrm{Z}^\mathcal V),\mathrm{Z}^\mathcal A\sim h_{\phi_2}(\mathrm{Z}^\mathcal A)}}\log p(\mathrm X|\mathrm{Z}^\mathcal V,\mathrm{Z}^\mathcal A)\\& -\text{KL}(h_{\phi_1}(\mathrm{Z}^\mathcal V)||p(\mathrm{Z}^\mathcal V))-\text{KL}(h_{\phi_2}(\mathrm{Z}^\mathcal A)||p(\mathrm{Z}^\mathcal A))\\
      &=\underline{\mathcal{L}}.
  \end{aligned}
\end{equation}
where  $\underline{\mathcal L}$ is the evidence lower bound that satisfies $\log(\mathrm A,\mathrm X)\ge\underline{\mathcal L}$,  $\text{KL}(p(\cdot)\ ||\ q(\cdot))$ is the Kullback-Leibler deivergence that compare the difference between probability distribution $p$ and $q$, $h_{\phi_1}(\mathrm Z^\mathcal V)$ and $h_{\phi_2}(\mathrm Z^\mathcal A)$ are variational posterior distributions generated from the node encoder and the attribute encoder respectively. By sampling latent embeddings $\mathrm Z^V$ and $\mathrm Z^A$  from  $h_{\phi_1}(\mathrm Z^V)$ and $h_{\phi_2}(\mathrm Z^A)$ and inputing  into decoder , $ p(\mathrm A|\mathrm Z^\mathcal V)$ and $p(\mathrm X|\mathrm Z^\mathcal V,\mathrm Z^\mathcal A)$ obtained by decoder should be close to observed data .
Generally, 
to maximize $\underline {\mathcal L}$, we need to consider the reconstruction loss and the KL divergence.

\section{Additional Experiments}
\label{sec:app:exp}

\subsection{Memory cost}
To compare the memory cost, 
we further evaluate the results of \name, HGCA and SeeGera.
We set the hidden space dimensionality as 64 in our experiments and give the results in Table~\ref{memo}. From the table, we see that GraMI has the smallest memory cost compared with others.

\begin{table}[t]
\centering
\caption{The Memory cost experiments on three datasets.}
\resizebox{0.7\linewidth}{!}
{
\begin{tabular}{c|ccc}
\hline
Dataset & SeeGera & HGCA & GraMI\\
\hline
ACM & 6697MB & 5314MB & 4413MB\\
DBLP & 21348MB & 19415MB & 11231MB\\
YELP & 4736MB & 3013MB & 1925MB\\ 
\hline
\end{tabular}
}
\label{memo}
\end{table}

\subsection{Sensitivity analysis on hidden embedding dimensionality}
\begin{table}[t]
\centering
\caption{The performance results (\%) w.r.t. the hidden node embedding dimensionality.}
\resizebox{\linewidth}{!}{
\begin{tabular}{c||c|c|cccc}
\hline
{Dataset} & {Metric} & Train & 32  & 64 & 128 & 256\\
\hline
\multirow{10}{*}{ACM} & \multirow{5}{*}{} & 10\% &  90.25 &91.07 & 91.58 &90.86\\
\multirow{10}{*}{} & Macro & 20\% & 90.70 & 92.03 & 92.12 & 91.63\\
\multirow{10}{*}{} & -F1 & 40\% & 91.03 & 92.52 & 92.89 & 92.04\\
\multirow{10}{*}{} & \multirow{5}{*}{} & 60\% &  91.40& 93.35 & 93.41 &92.36\\
\multirow{10}{*}{} & \multirow{5}{*}{} & 80\% &91.69 & 93.44 & 93.50&92.84\\
\cline{2-7}
\multirow{10}{*}{} & \multirow{5}{*}{} & 10\% & 90.15 & 91.07 & 91.47& 90.78\\
\multirow{10}{*}{} & Micro & 20\% &90.58 & 91.92 & 91.98 & 91.51 \\
\multirow{10}{*}{} & -F1 & 40\% & 90.89 & 92.50 & 92.77 & 91.91\\
\multirow{10}{*}{} & \multirow{5}{*}{} & 60\% & 91.37 & 93.25 & 93.27 & 92.24\\
\multirow{10}{*}{} & \multirow{5}{*}{} & 80\% & 91.59 & 93.32 & 93.38 & 92.76\\
\hline
\multirow{10}{*}{DBLP} & \multirow{5}{*}{} & 10\% & 93.59 & 93.51 & 93.89 & 94.11\\
\multirow{10}{*}{} & Macro & 20\% & 93.61 & 93.82 & 94.14 & 94.30 \\
\multirow{10}{*}{} & -F1 & 40\% &93.72 & 93.85 & 94.30  & 94.31
\\
\multirow{10}{*}{} & \multirow{5}{*}{} & 60\% & 93.73 & 93.81 & 94.29 & 94.82\\
\multirow{10}{*}{} & \multirow{5}{*}{} & 80\%& 93.65 & 93.24 & 94.02 & 94.64\\
\cline{2-7}
\multirow{10}{*}{} & \multirow{5}{*}{} & 10\% & 94.02 & 94.12 & 94.27 & 94.56\\
\multirow{10}{*}{} & Micro & 20\% & 94.03 & 94.23 & 94.5  & 94.73\\
\multirow{10}{*}{} & -F1 & 40\% & 94.15 & 94.27 & 94.67 & 94.75 
\\
\multirow{10}{*}{} & \multirow{5}{*}{} & 60\% & 94.17 & 94.22 & 94.64 & 95.18 \\
\multirow{10}{*}{} & \multirow{5}{*}{} & 80\% & 94.17 & 94.14 & 94.44 & 95.06\\
 \hline
\multirow{10}{*}{YELP} & \multirow{5}{*}{} & 10\% & 90.45 & 91.03 & 89.82 & 88.14\\
\multirow{10}{*}{} & Macro & 20\% & 90.76 & 91.15 & 90.06 & 88.76\\
\multirow{10}{*}{} & -F1 & 40\% &91.87 & 91.99 & 91.54  & 90.14
\\
\multirow{10}{*}{} & \multirow{5}{*}{} & 60\% & 92.70 & 92.84 & 92.44 & 90.58\\
\multirow{10}{*}{} & \multirow{5}{*}{} & 80\%& 93.34 & 93.18 & 93.39 & 91.34 \\
\cline{2-7}
\multirow{10}{*}{} & \multirow{5}{*}{} & 10\% & 90.52 & 90.76 & 89.61 & 88.91\\
\multirow{10}{*}{} & Micro & 20\% & 90.88 & 90.87 & 89.97  & 89.58\\
\multirow{10}{*}{} & -F1 & 40\% & 91.23 & 91.56 & 91.25 & 90.18
\\
\multirow{10}{*}{} & \multirow{5}{*}{} & 60\% & 92.13 & 92.36 & 92.09 & 90.77\\
\multirow{10}{*}{} & \multirow{5}{*}{} & 80\% & 92.68 & 93.14 & 93.13 & 91.49 \\
\hline
\end{tabular}
}
\label{latent}
\end{table}
We next conduct experiments on the hyperparameter sensitivity analysis on hidden embedding dimensionality. 
We vary the dimensionality from \{32, 64, 128, 256\}.
The results are presented in Table~\ref{latent}. 
We see that \name\ is insensitive to different dimensionality sizes. 

\subsection{HGNN model}
To validate the generality of the HGNN model used in \name, 
we further replace the attention-based HGNN with RGCN~\cite{schlichtkrull2018modeling}. 
The experimental results are shown in Table~\ref{table:rgcn}.
We see that,
with RGCN as the HGNN, \name\ can also achieve decent performance, which verifies that
GraMI is suitable for other HGNN models.

\begin{table}[!htbp]
\centering
\caption{The results(\%) with RGCN model on three datasets.}

{
\begin{tabular}{c|c|c|c|c}
\hline
{Metric} & Training & ACM  & DBLP & YELP\\
\hline
\multirow{5}{*}{} & 10\% &  90.07 &92.40 & 89.30\\
Macro & 20\% & 90.51 & 93.10 & 89.72\\
-F1 & 40\% & 91.15 & 93.24 & 90.42\\
\multirow{5}{*}{} & 60\% &  91.70& 93.35 & 90.83\\
\multirow{5}{*}{} & 80\% &92.48 & 93.53 & 92.11\\
\hline
\multirow{5}{*}{} & 10\% & 89.92 & 92.93 &89.52\\
 Micro & 20\% &90.36 & 93.55 &89.81\\
 -F1 & 40\% & 90.91 & 93.70 &90.48\\
 \multirow{5}{*}{} & 60\% & 91.54 & 93.74 & 90.75\\
 \multirow{5}{*}{} & 80\% & 92.36 & 94.03 & 92.89\\
\hline

\end{tabular}
}
\label{table:rgcn}
\end{table}

\subsection{Quality of Generated Attributes}
\label{sec:app:quality}
The full results on evaluating the quality of generated attributes are given in Table~\ref{table:app:quality}.
From the table, we see that \name\ can generate high-quality attributes that can lead to better model performance across various noise rates.

\begin{table}[t]
\centering
\caption{Node classification results (\%) on MAGNN.}
\resizebox{0.95\linewidth}{!}
{
\begin{tabular}{c||c|c|cccc}
\hline
\multirow{2}{*}{Dataset} & \multirow{2}{*}{Metric} & Train & MAGNN  & MAGNN & MAGNN & MAGNN\\
\multirow{2}{*}{} & \multirow{2}{*}{} & -ing & -AVG  & -onehot & -AC & -\name\\
\hline
\multirow{10}{*}{ACM} & \multirow{5}{*}{} & 10\% & 88.82&89.69	&{92.92}&\textbf{93.34}*\\
\multirow{10}{*}{} & Macro & 20\% & 89.41&	90.61&	{93.34}	&\textbf{94.98}**\\
\multirow{10}{*}{} & -F1 & 40\% & 89.83&	92.48&	{93.72}&	\textbf{94.35}**\\
\multirow{10}{*}{} & \multirow{5}{*}{} & 60\% & 90.18&	93.12&	{94.01}&	\textbf{94.98}*
 \\
\multirow{10}{*}{} & \multirow{5}{*}{} & 80\% &90.11	&93.20&	{94.08}&	\textbf{94.76}*
 \\
\cline{2-7}
\multirow{10}{*}{} & \multirow{5}{*}{} & 10\% & 88.81&89.92	&{92.33}&\textbf{93.52}**\\
\multirow{10}{*}{} & Micro & 20\% & 89.36	&90.58	&{93.21}&	\textbf{94.16}**
 \\
\multirow{10}{*}{} & -F1 & 40\% & 89.81&	92.93&	{93.60}&	\textbf{94.49}**
 \\
\multirow{10}{*}{} & \multirow{5}{*}{} & 60\% & 90.11&	93.52&	{93.87}&	\textbf{95.13}**
 \\
\multirow{10}{*}{} & \multirow{5}{*}{} & 80\% & 90.06	&93.65&{93.93}&	\textbf{94.93}**
 \\
\hline
\multirow{10}{*}{DBLP} & \multirow{5}{*}{} & 10\% & 92.52&92.64	&{94.01}&\textbf{94.21}
\\
\multirow{10}{*}{} & Macro & 20\% & 92.70&	92.73&	{94.16}	&\textbf{94.47}*
\\
\multirow{10}{*}{} & -F1 & 40\% &92.69	&93.19&	{94.29}	&\textbf{94.55}
\\
\multirow{10}{*}{} & \multirow{5}{*}{} & 60\% & 92.75&	93.54&	{94.35}&	\textbf{94.64}*
\\
\multirow{10}{*}{} & \multirow{5}{*}{} & 80\% & 93.01&	94.01&	\textbf{94.53}	&{94.50}
 \\
\cline{2-7}
\multirow{10}{*}{} & \multirow{5}{*}{} & 10\% & 93.08&93.25&{94.26}&\textbf{94.61}*	
 \\
\multirow{10}{*}{} & Micro & 20\% & 93.25&	93.27&	{94.58}&	\textbf{94.85}*
 \\
\multirow{10}{*}{} & -F1 & 40\% & 93.25&	93.69&	{94.71}&	\textbf{94.92}
\\
\multirow{10}{*}{} & \multirow{5}{*}{} & 60\% & 93.34&	94.03&	{94.77}&	\textbf{95.00}
 \\
\multirow{10}{*}{} & \multirow{5}{*}{} & 80\% & 93.57&	94.45&	\textbf{94.92}&	{94.90}
 \\
\hline
\multirow{10}{*}{YELP} & \multirow{5}{*}{} & 10\% &86.86&87.09&{89.54}&\textbf{91.04}**
\\
\multirow{10}{*}{} & Macro & 20\% & 87.86&	88.59	&{89.63}	&\textbf{91.30}**
\\
\multirow{10}{*}{} & -F1 & 40\% & 89.85&	90.50	&{91.32}	&\textbf{91.60}
 \\
\multirow{10}{*}{} & \multirow{5}{*}{} & 60\% & 90.58	&91.41	&\textbf{91.67}	&{91.44}
\\
\multirow{10}{*}{} & \multirow{5}{*}{} & 80\% & 90.57	&91.30	&{91.63}	&\textbf{91.98}
 \\
\cline{2-7}
\multirow{10}{*}{} & \multirow{5}{*}{} & 10\% &86.68&87.20&{89.03}&\textbf{90.58}**
 \\
\multirow{10}{*}{} & Micro & 20\% &87.84&	88.67&	{89.42}&	\textbf{90.82}**
 \\
\multirow{10}{*}{} & -F1 & 40\% & 89.86&	90.41&	{90.99}&	\textbf{91.08}
 \\
\multirow{10}{*}{} & \multirow{5}{*}{} & 60\% & 90.64	&{91.26}	&\textbf{91.36}&	{90.97}
\\
\multirow{10}{*}{} & \multirow{5}{*}{} & 80\% & 90.62&	91.19&	{91.35}	&\textbf{91.49}
\\
\hline
\end{tabular}
}
\label{table:app:quality}
\end{table}

\subsection{Experiments on million-scale graphs}
We further conduct experiments on the ogbn-mag dataset~\cite{hu2020open} with roughly 2 million nodes. It contains four types of entities: papers (736,389 nodes),
authors (1,134,649 nodes), institutions (8,740 nodes), and fields of study (59,965 nodes).
We show node classification accuracy in Tabel~\ref{ogb}. 
We see that GraMI can run on the large-scale dataset and performs well under different batch sizes.
Compared with the HGNN model R-GCN, 
\name\ can complete missing attributes and correct noisy ones, which leads to better performance.
\begin{table}[!hbpt]
\centering
\caption{Accuracy (\%) on node classification for ogbn-mag.}
{
\begin{tabular}{c|c|c}
\hline
Batch size & RGCN &  \name \\
\hline
1024 & $47.32 \pm {0.21}$ & $\textbf{54.68}\pm \textbf{0.15}$ \\
2048 & ${48.06}\pm {0.14}$ & $\textbf{56.73}\pm \textbf{0.23}$\\
\hline
\end{tabular}
}
\label{ogb}
\end{table}


\end{document}